\documentclass[10pt,twocolumn,letterpaper]{article}

\usepackage[pagenumbers]{cvpr} 

\usepackage{graphicx}
\usepackage{amsmath}
\usepackage{amssymb}
\usepackage{booktabs}

\usepackage[pagebackref,breaklinks,colorlinks]{hyperref}
\usepackage{multirow}
\usepackage{subcaption}
\usepackage{bbm}
\usepackage[linesnumbered, algoruled]{algorithm2e}
\usepackage{caption} 
\usepackage{enumitem}

\usepackage[capitalize]{cleveref}
\crefname{section}{Sec.}{Secs.}
\Crefname{section}{Section}{Sections}
\Crefname{table}{Table}{Tables}
\crefname{table}{Tab.}{Tabs.}

\def\cvprPaperID{5099} 
\def\confName{CVPR}
\def\confYear{2022}

\begin{document}
	
	\title{SoftGroup for 3D Instance Segmentation on Point Clouds}
	
	\author{Thang Vu \quad Kookhoi Kim \quad Tung M. Luu \quad Xuan Thanh Nguyen \quad Chang D. Yoo\\
		Korea Advanced Institute of Science and Technology (KAIST)\\
		{\tt\small \{thangvubk,rlarnrghlapz,tungluu2203,thanhnguyen,cd\_yoo\}@kaist.ac.kr}
	}
	\maketitle
	
	\begin{abstract}
		Existing state-of-the-art 3D instance segmentation methods perform semantic segmentation followed by grouping. The hard predictions are made when performing semantic segmentation such that each point is associated with a single class. However, the errors stemming from hard decision propagate into grouping that results in (1) low overlaps between the predicted instance with the ground truth and (2) substantial false positives. To address the aforementioned problems, this paper proposes a 3D instance segmentation method referred to as SoftGroup by performing bottom-up soft grouping followed by top-down refinement. SoftGroup allows each point to be associated with multiple classes to mitigate the problems stemming from semantic prediction errors and suppresses false positive instances by learning to categorize them as background. Experimental results on different datasets and multiple evaluation metrics demonstrate the efficacy of SoftGroup. Its performance surpasses the strongest prior method by a significant margin of +6.2\% on the ScanNet v2 hidden test set and +6.8\% on S3DIS Area 5 in terms of AP$_{50}$. SoftGroup is also fast, running at 345ms per scan with a single Titan X on ScanNet v2 dataset. The source code and trained models for both datasets are available at \url{https://github.com/thangvubk/SoftGroup.git}.

	\end{abstract}
	
	\section{Introduction}
	\label{sec:intro}
	
	\begin{figure}
		\includegraphics[width=\columnwidth]{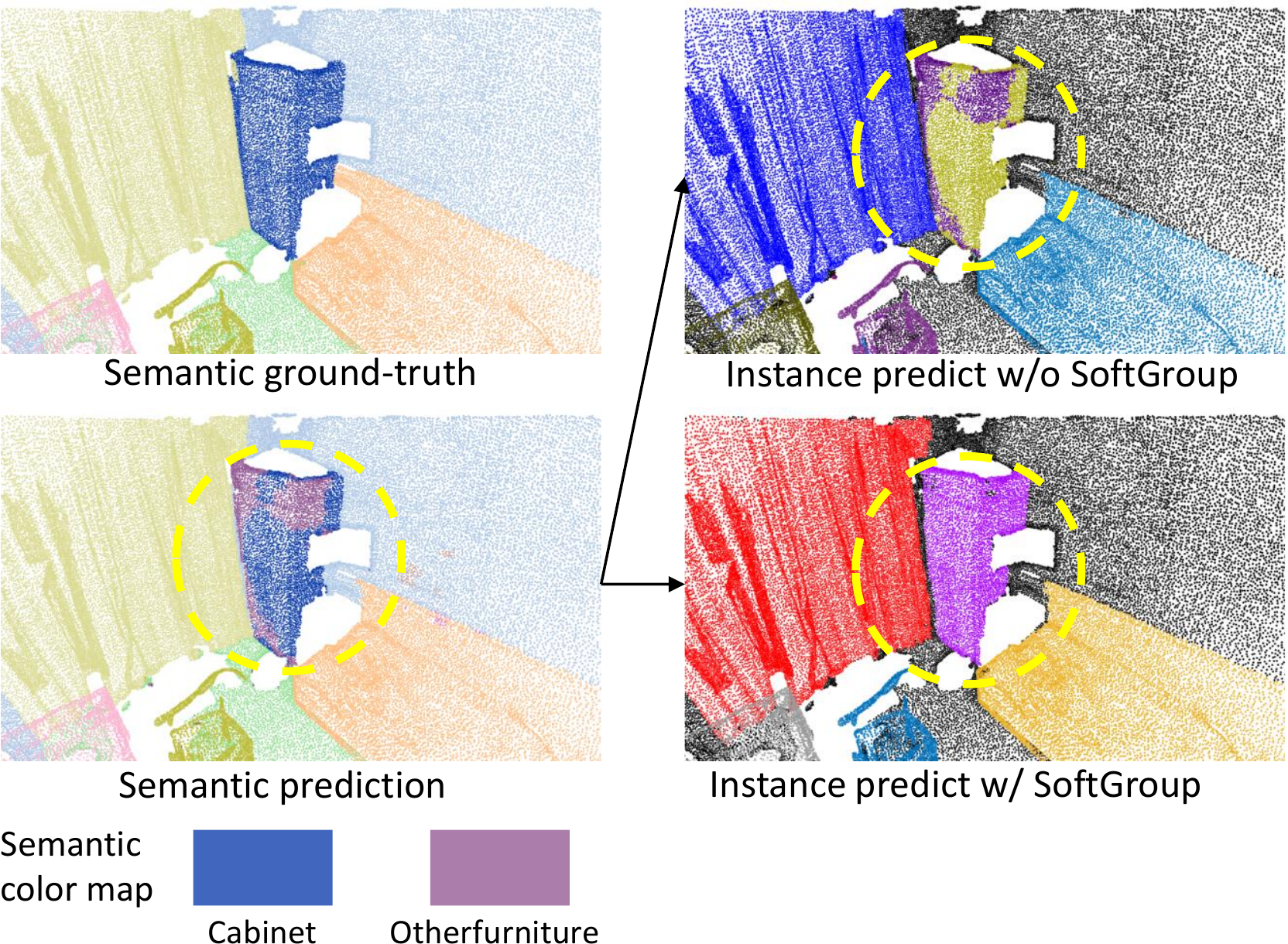}
		\caption{Instance segmentation with and without SoftGroup from the same semantic prediction results. The last row shows the palette for semantic predictions only. Instance predictions are illustrated by different random colors for different objects. In the semantic prediction results, some regions of \texttt{cabinet} are wrongly predicted as \texttt{other furniture}. Without SoftGroup, these errors are propagated to instance prediction. SoftGroup addresses this problem and produces more accurate instance masks.}
		\label{fig:introduction}
	\end{figure}
	
	Scene understanding on 3D data has received increasing attention for the rapid development of 3D sensors and availability of large-scale 3D datasets. Instance segmentation on point clouds is a 3D perception task, serving as the foundation for a wide range of applications such as autonomous driving, virtual reality, and robot navigation. Instance segmentation processes the point clouds to output a category and an instance mask for each detected object. 
	
	State-of-the-art methods \cite{jiang2020pointgroup,liang2021instance,chen2021hierarchical} consider 3D instance segmentation as a bottom-up pipeline. They learn the point-wise semantic labels and center offset vectors and then group points of the same labels with small geometric distances into instances. These grouping algorithms are performed on the hard semantic prediction, where a point is associated with a single class. In many cases, objects are locally ambiguous, the output semantic predictions show different categories for different parts, and thus using hard semantic predictions for instance grouping leads to two problems: (1) low overlap between predicted instance and the ground-truth and (2) extra false-positive instances from wrong semantic regions. \Cref{fig:introduction} shows a visualization example. Here, in the semantic prediction results, some parts of \texttt{cabinet} is wrongly predicted as \texttt{other furniture}. When hard semantic predictions are used to perform grouping, the semantic prediction error is propagated to instance prediction. As a result, the predicted \texttt{cabinet} instance has low overlap with the ground truth, and the \texttt{other furniture} instance is a false positive. 
	
\begin{figure}
		\centering
		\includegraphics[width=\columnwidth]{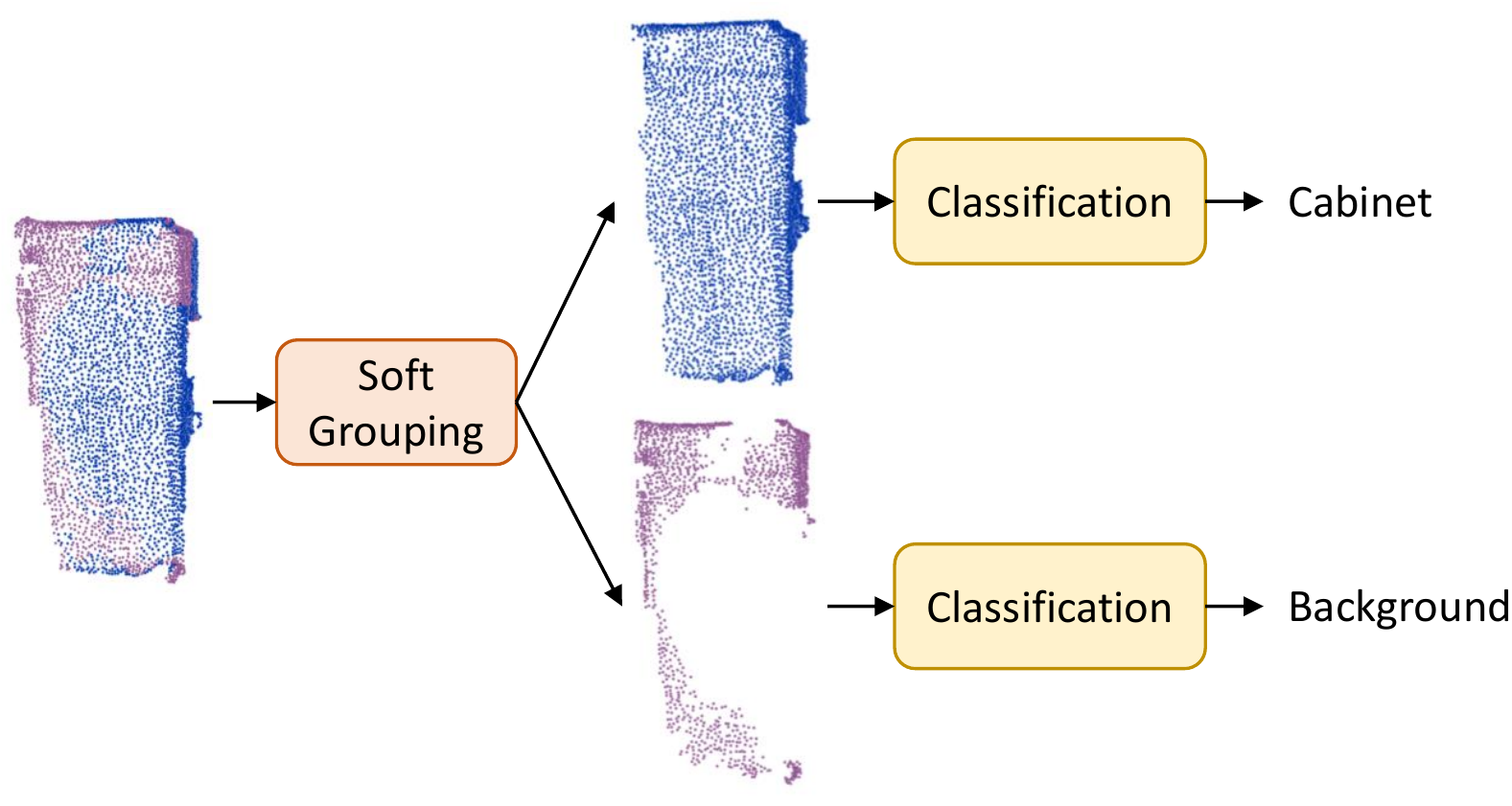}
		\caption{The \texttt{cabinet} in \Cref{fig:introduction} is extracted to illustrate the high-level pipeline of our method. The soft grouping module based on soft semantic scores to output more accurate instance (the upper one). The classifier processes each instance and suppress the instance from wrong semantic prediction (the lower one).}
		\label{fig:intuation}
	\end{figure}

	This paper proposes SoftGroup to address these problems by considering soft semantic scores to perform grouping instead of hard one-hot semantic predictions. The intuition of SoftGroup is illustrated in \Cref{fig:intuation}. Our finding is that the object parts with wrong semantic predictions still have reasonable scores for the true semantic class. SoftGroup relies on a score threshold to determine which category the object belongs instead of the argument max values. Grouping on the soft semantic scores produces for accurate instance on true semantic class. The instance with wrong semantic prediction will be suppressed by learning to categorize it as background. To this end, we treat an instance proposal as either a positive or negative sample depending on the maximum Intersection over Union (IoU) with the ground truth, then construct a top-down refinement stage to refine the positive sample and suppress the negative one. As shown in \Cref{fig:introduction}, SoftGroup is able to produce accurate instance masks from imperfect semantic prediction. 
	
	SoftGroup is conceptually simple and easy to implement. Experiments on the ScanNet v2 \cite{dai2017scannet} and S3DIS \cite{armeni20163d} benchmark datasets show the efficacy of our method. Notably, SoftGroup outperforms the previous state-of-the-art method by a significant margin of +6.2\% on the ScanNet hidden test set and +6.8\% on S3DIS Area 5 in terms of AP$_{50}$. SoftGroup is fast, requiring 345ms to process a ScanNet scene. In summary, our contribution is threefold.
	\begin{itemize}[topsep=4pt]
		\item We propose SoftGroup that performs grouping on soft semantic scores to address the problem of the hard semantic predictions that propagates the errors to instance predictions.
		\item We propose a top-down refinement stage to correct, refine the positive samples and suppress false positives introduced by wrong semantic predictions.
		\item We report extensive experiments on multiple datasets with different evaluation metrics, showing significant improvements over existing state-of-the-art methods.
	\end{itemize}
	
	%
	
	\begin{figure*}
		\centering
		\includegraphics[width=\textwidth]{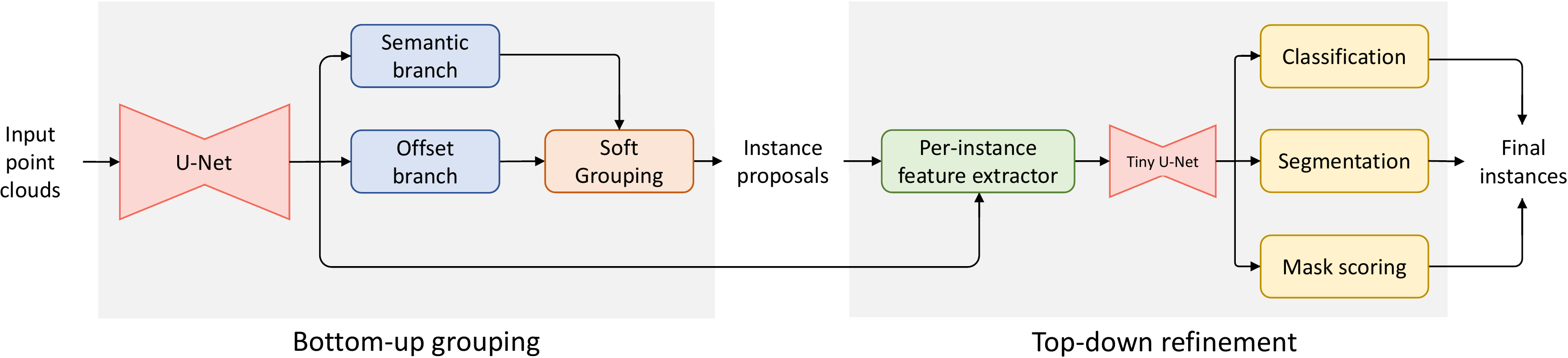}
		\caption{The architecture of the proposed method consists of bottom-up grouping and top-down refinement stages. From the input point clouds, the U-Net backbone extracts the point features. Then semantic and offset branches predict the semantic scores and offset vectors, followed by a soft grouping module to generate instance proposal. The feature extractor layer extracts backbone features from instance proposals. The features for each proposal are fed into a tiny U-Net followed by the classification, segmentation, and mask scoring branches to get the final instances.}
		\label{fig:arch}
	\end{figure*}
	
	\section{Related work}
	\paragraph{Deep Learning on 3D Point Clouds. }
	Point cloud representation is a common data format for 3D scene understanding since it is simple while preserving original geometric information. To process point clouds, early methods \cite{aubry2011wave,rusu2009fast,rusu2008aligning,bronstein2010scale} extract hand-crafted features based on statistical properties of points. Recent deep learning methods learn to extract features from points. Pointwise methods, such as PointNet \cite{qi2017pointnet,qi2017pointnet++}, directly process points through shared Multi-Layer Perceptron (MLP) and then aggregate regional and global features from symmetric function, such as max-pooling. Voxel-based methods \cite{graham20183d,choy20194d,maturana2015voxnet,Riegler2017OctNet} transform the unordered point sets into ordered sparse volumetric grids and then perform 3D sparse convolutions on the grids, showing the effectiveness in performance and speed.

	\paragraph{Proposal-based Instance Segmentation.}
	Proposal-based methods consider a top-down strategy that generates region proposals and then segments the object within each proposal. Existing proposal-based methods for 3D point clouds are highly influenced by the success of Mask-R CNN for 2D images. To handle data irregularity of point clouds,  Li \etal \cite{yi2019gspn} propose GSPN, which takes an analysis-by-synthesis strategy to generate high-objectness 3D proposals, which are refined by a region-based PointNet. Hou \etal \cite{hou20193d} present 3DSIS that combines multi-view RGB input with 3D geometry to predict bounding boxes and instance masks. Yang \etal \cite{yang2019learning} propose 3D-BoNet which directly outputs a set of bounding boxes without anchor generation and non-maximum suppression, then segments the object by a pointwise binary classifier. Liu \etal \cite{liu2020learning} present GICN to approximate the instance center of each object as a Gaussian distribution, which is sampled to get object candidates then produce corresponding bounding boxes and instance masks.

	\paragraph{Grouping-based Instance Segmentation.}
	Grouping-based methods rely on a bottom-up pipeline that produces per-point predictions (such as semantic maps, and geometric shifts, or latent features) then groups points into instances. Wang \etal \cite{wang2018sgpn} propose SGPN to construct a feature similarity matrix for all points and then group points of similar features into instances. Pham \etal \cite{pham2019jsis3d} present JSIS3D that incorporates the semantic and instance labels by a multi-value conditional random field model and jointly optimizes the labels to obtain object instances. Lahoud \etal \cite{lahoud20193d} propose MTML to learn feature and directional embedding, then perform mean-shift clustering on the feature embedding to generate object segments which are scored according to their direction feature consistency. Han \etal \cite{han2020occuseg} introduce OccuSeg that performs graph-based clustering guided by object occupancy signal for more accurate segmentation outputs. Zhang \etal \cite{zhang2021point} consider a probabilistic approach that represents each point as a tri-variate normal distribution followed by a clustering step to obtain object instances. Jiang \etal \cite{jiang2020pointgroup} propose PointGroup to segment objects on original and offset-shifted point sets, relying on a simple yet effective algorithm that groups nearby points of the same label and expands the group progressively. Chen \etal \cite{chen2021hierarchical} extend PointGroup and propose HAIS that further absorbs surrounding fragments of instances and then refines the instances based on intra-instance prediction. Liang \etal \cite{liang2021instance} SSTNet to construct a tree network from pre-computed superpoints then traverse the tree and split nodes to get object instances.
	
	The common proposal-based and grouping-based methods have their advantages and drawbacks. Proposal-based methods process each object proposal independently that is not interfered with by other instances. Grouping-based methods process the whole scene without proposal generation, enabling fast inference. However, proposal-based methods have difficulties in generating high-quality proposals since the point only exists on the object surface. Grouping-based methods highly depend on semantic segmentation such that the errors in semantic predictions are propagated to instance predictions. The proposed method leverages the advantages and address the limitations of both approaches. Our method is constructed as a two-stage pipeline, where the bottom-up stage generates high-quality object proposals by grouping on soft semantic scores, and then the top-down stage process each proposal to refine positive samples and suppress negative ones.
	
	\section{Method}

	The overall architecture of SoftGroup is depicted in Figure \ref{fig:arch}, which is divided into two stages. In the bottom-up grouping stage, the point-wise prediction network (Sec. \ref{ssec:point_wise_net}) takes point clouds the input and produces point-wise semantic labels and offset vectors. The soft grouping module (Sec. \ref{ssec:softgroup}) processes these outputs to produce preliminary instance proposals. In the top-down refinement stage, based on the proposals, the corresponding features from the backbone are extracted and used to predict classes, instance masks, and mask scores as the final results.
	
	\subsection{Point-wise Prediction Network}
	\label{ssec:point_wise_net}
	The input of the point-wise prediction network is a set of $N$ points, each of which is represented by its coordinate and color. The point set is voxelized to convert unordered points to ordered volumetric grids, which are fed into a U-Net style backbone \cite{ronneberger2015u} to obtain point features. The Submanifold Sparse Convolution \cite{graham20183d} is adopted to implement the U-Net for 3D point clouds. From the point features, two branches are constructed to output the point-wise semantic scores and offset vectors.
	
	\paragraph{Semantic Branch.} A semantic branch is constructed from a two-layer MLP and learns to output semantic scores $\boldsymbol{S} = \{\boldsymbol{s}_1, ..., \boldsymbol{s}_N\} \in \mathbb{R}^{N\times N_\text{class}}$ for $N$ points over $N_\text{class}$ classes. Different from existing methods \cite{jiang2020pointgroup,chen2021hierarchical}, we directly perform grouping on semantic scores without converting the semantic scores to one-hot semantic predictions. 
	
	\paragraph{Offset Branch.} In parallel with the semantic branch, we apply a two-layer MLP to learn the offset vectors $\boldsymbol{O} = \{\boldsymbol{o}_1, ..., \boldsymbol{o}_N\} \in \mathbb{R}^{N\times 3}$, which represents the vector from each point to the geometric center of the instance the point belongs. Based on the learned offset vectors, we shift the points to the center of the corresponding instance to perform grouping more effectively.
	
	The cross-entropy loss and $\ell_1$ regression loss are used to train the semantic and offset branches, respectively.
	
	\begin{equation}
	L_{\text{semantic}} = \frac{1}{N} \sum_{i=1}^{N}\text{CE}(\boldsymbol{s}_i, {s}^*_i),
	\end{equation}
	\begin{equation}
	L_{\text{offset}} = \frac{1}{\sum_{i=1}^{N} \mathbbm{1}_{\{\boldsymbol{p}_i\}}} \sum_{i=1}^{N} \mathbbm{1}_{\{\boldsymbol{p}_i\}} \Vert \boldsymbol{o}_i - \boldsymbol{o}_i^* \Vert_1,
	\end{equation}
	where ${s}^*$ is the semantic label, $\boldsymbol{o}^*$ is offset label representing the vector from a point to the geometric center of the instance that the point belongs to (analogous to \cite{jiang2020pointgroup,chen2021hierarchical,liang2021instance}), and $\mathbbm{1}_{\{\boldsymbol{p}_i\}}$ is the indicator function indicating whether the point $\boldsymbol{p}_i$ belongs to any instance.
	\subsection{Soft Grouping}
	\label{ssec:softgroup}
	The soft grouping module receives the semantic scores and offset vectors as the input and produces instance proposals. First, the offset vectors are used to shift points toward the corresponding instance centers. To perform grouping using the semantic scores, we define a score threshold $\tau$ to determine which semantic classes a point belongs to, allowing the point to be associated with multiple classes. Given semantic scores $\boldsymbol{S} \in \mathbb{R}^{N\times N_\text{class}}$, we iterate through $N_\text{class}$ classes, and at each class index we slice a point subset of the whole scene that has the score (w.r.t. the class index) higher than the threshold $\tau$. We follow \cite{jiang2020pointgroup,chen2021hierarchical} to perform grouping on each point subset. Since all points in each subset belong to the same class, we simply traverse all the points in the subset and create the links between points having a geometric distance smaller than a grouping bandwidth $b$ to get the instance proposals. For each iteration, the grouping is performed on a point subset of the whole scan, ensuring fast inference. The overall instance proposals are the union of the proposals from all subsets.
	
	We note that existing proposal-based methods \cite{hou20193d,yang2019learning,liu2020learning} commonly consider bounding boxes as object proposals then perform segmentation within each proposal. Intuitively, the bounding box with high overlap with the instance should have the center close to the object center. However, generating high-quality bounding box proposals in 3D point clouds is challenging since the point only exists on object surfaces. Instead, SoftGroup relies on point-level proposals which are more accurate and naturally inherit the scattered property of point clouds. 
	
	\begin{figure}
		\centering
		\begin{subfigure}{0.495\columnwidth}
			\includegraphics[width=\textwidth]{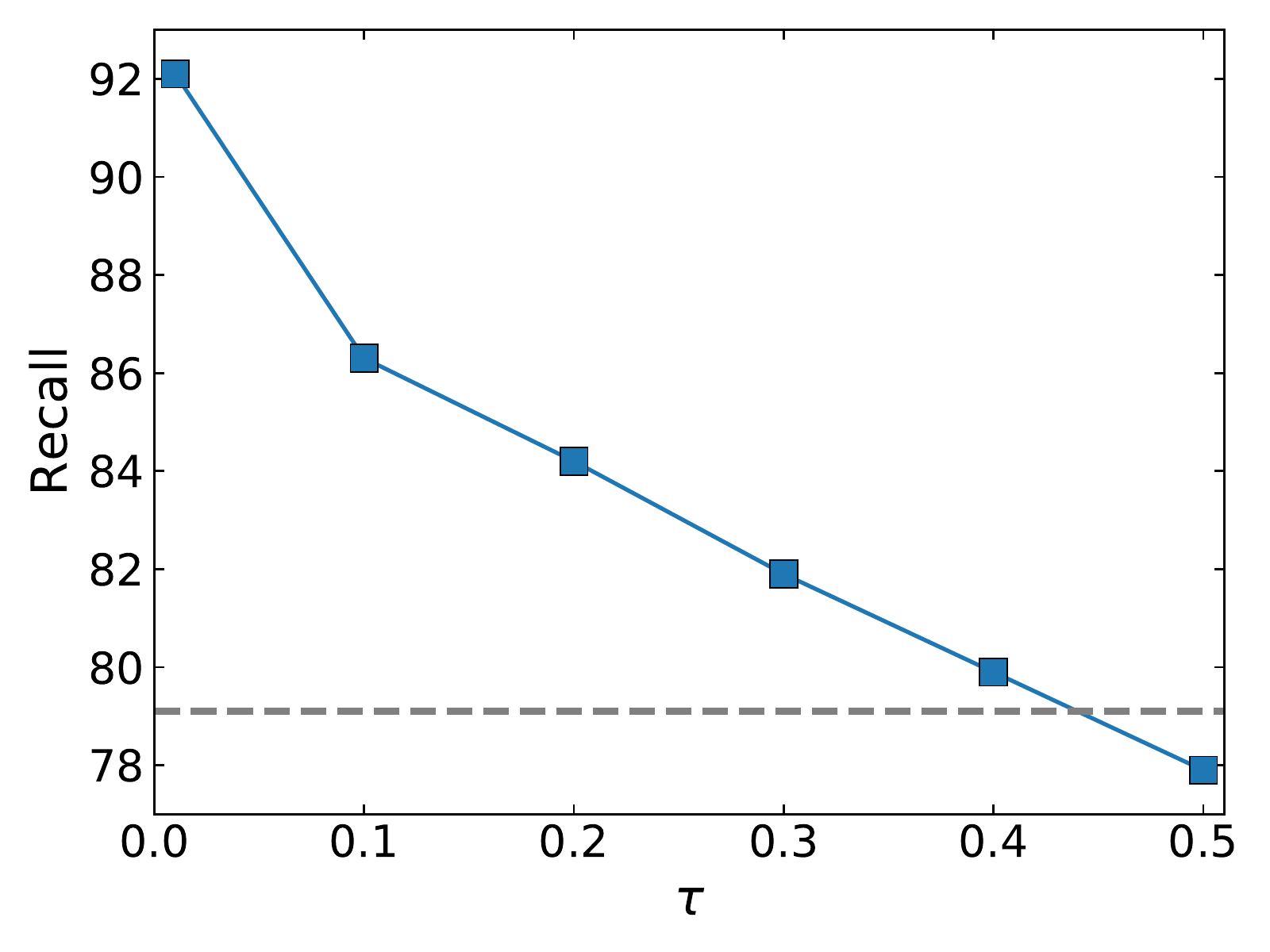}
		\end{subfigure}
		\begin{subfigure}{0.495\columnwidth}
			\includegraphics[width=\textwidth]{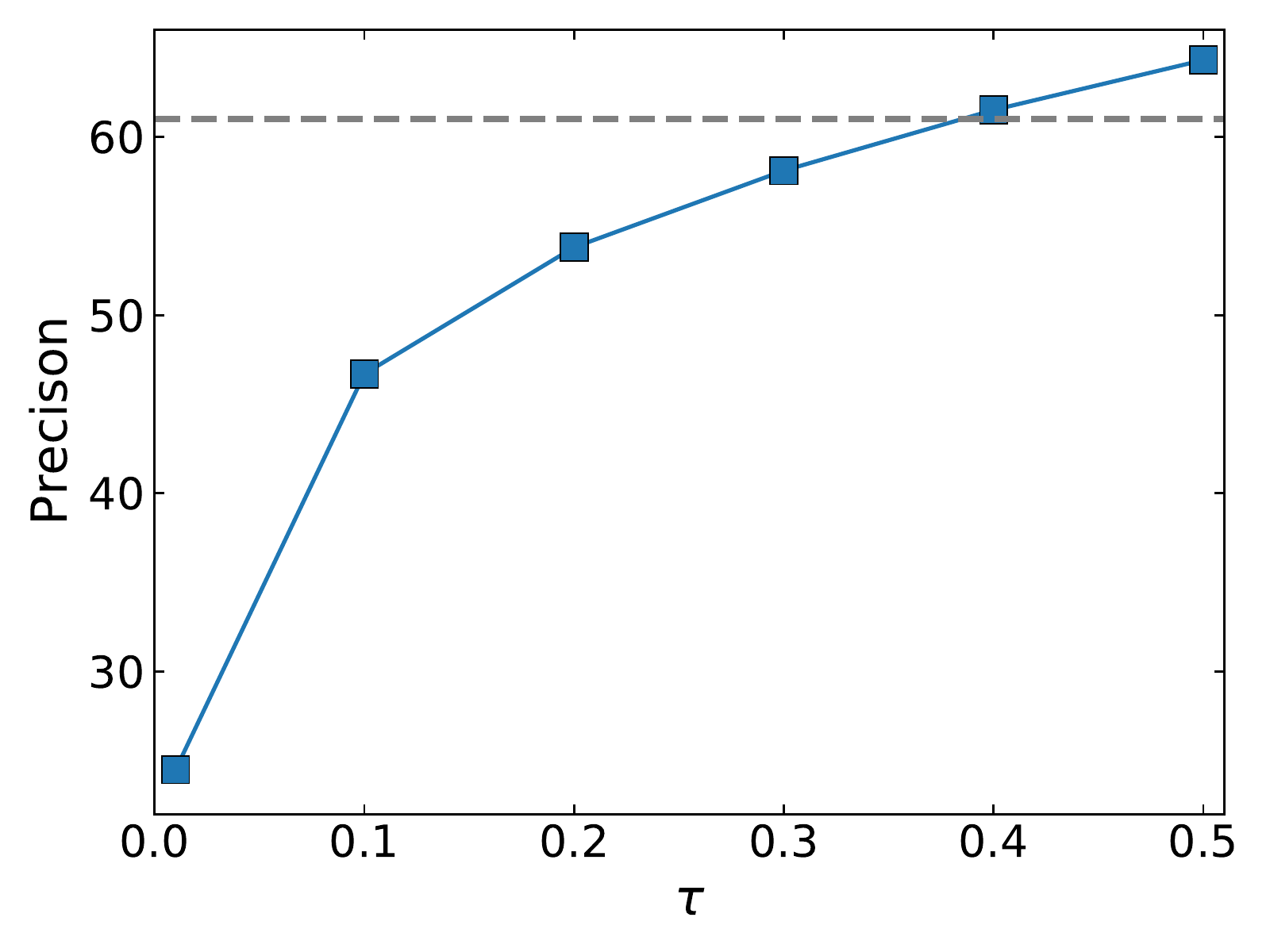}
		\end{subfigure}
		\caption{The recall and precision of semantic prediction with varying score threshold $\tau$. The dashed lines denote the recall and precision with hard semantic prediction.}
		\label{fig:recall_precision}
	\end{figure}
	
	Since the quality of instance proposals from grouping highly depend on the quality of semantic segmentation, we quantitatively analyze the impact of $\tau$ on the recall and precision of semantic predictions. The recall and precision for class $j$ is defined as follows.
	\begin{equation}
	\begin{aligned}
	\text{recall}_j &= \sum_{i=1}^{N} \frac{(s_{ij} > \tau) \land (s^*_i = j)}{{s^*_i = j}},\\
	\text{precision}_j &= \sum_{i=1}^{N} \frac{(s_{ij} > \tau) \land (s^*_i = j)}{s_{ij} > \tau}.
	\end{aligned} 
	\end{equation}
	
	\Cref{fig:recall_precision} shows the recall and precision (averaged over classes) with the varying score thresholds $\tau$ compared with those of hard semantic prediction. With hard semantic prediction, the recall is 79.1\%, indicating more than 20\% amount of points over classes are not covered by the predictions. When using the score threshold, the recall increases as the score threshold decreases. However, the small score threshold also leads to low precision. We propose a top-down refinement stage mitigate the low precision problems. The precision can be interpreted as the relation between foreground and background points of object instances. We set the threshold to 0.2 with precision near 50\%, leading to the ratio between foreground and background points for ensuring stage is balanced.

	\subsection{Top-Down Refinement}
	\label{ssec:top_down_refinement}
	The top-down refinement stage classifies and refines the instance proposals from the bottom-up grouping stage. A feature extractor layer processes each proposal to extract its corresponding backbone features. The extracted features are fed into a tiny U-Net network (a U-Net style network with a small number of layers) before predicting classification scores, instance masks, and mask scores at the ensuing branches.
	
	\paragraph{Classification Branch.} The classification branch starts with a global average pooling layer to aggregate the feature of all points in the instance, followed by a MLP to predict the classification scores $\boldsymbol{C} = \{\boldsymbol{c}_1, ..., \boldsymbol{c}_K\} \in \mathbb{R}^{K\times (N_\text{class}+1)}$, where $K$ is the number of instances. We directly derive the object category and classification confidence score from the output of the classification branch. 
	
	We note that existing grouping-based methods typically derive the object category from semantic predictions. However, instances may come from objects with noisy semantic predictions. The proposed method directly uses the output of the classification branch as the instance class. The classification branch aggregates all point features of the instance and classifies the instance with a single label, leading to more reliable predictions. 
	
	\paragraph{Segmentation Branch.}
	As shown in \Cref{ssec:softgroup}, the instance proposals contain both foreground and background points, we construct a segmentation branch to predict an instance mask within each proposal. The segmentation branch is a point-wise MLP of two layers that output an instance mask $\boldsymbol{m}_k$  for each instance $k$.
	
	\paragraph{Mask Scoring Branch.}
	The mask scoring branch shares the same structure as the classification branch. This branch outputs the mask scores $\boldsymbol{E} = \{\boldsymbol{e}_1, ..., \boldsymbol{e}_K\} \in \mathbb{R}^{K\times N_\text{class}}$, which estimate the IoU of a predicted mask with the ground truth. The mask score is combined with the classification score by multiplication to get the final confidence score. 
	
	\paragraph{Learning Targets.} Training the top-down refinement branches requires the target labels for each branch. To this end, we follow the logic in existing 2D object detection and segmentation methods. We treat all instance proposals having IoU with a ground-truth instance higher than 50\% as the positive samples and the rest as negatives. Every positive sample is assigned to a ground-truth instance with the highest IoU. The classification target of a positive sample is the category of the corresponding ground-truth instance. The total number of classes is $N_\text{class}$ + 1 ($N_\text{class}$ foreground classes and one background class). The segmentation and mask scoring branches are trained on positive samples only. The mask target of a positive sample is the mask of the assigned ground-truth instance. The mask score target is the IoU between the predicted mask and the ground truth. The training loss of these branches is the combination of cross-entropy, binary cross-entropy, and $\ell_2$ regression losses,  following \cite{he2017mask,huang2019mask}.
	
	\begin{equation}
	L_{\text{class}} = \frac{1}{K} \sum_{k=1}^{K}\text{CE}(\boldsymbol{c}_k, {c}^*_k),
	\end{equation}
	\begin{equation}
	L_{\text{mask}} = \frac{1}{\sum_{k=1}^{K} \mathbbm{1}_{\{\boldsymbol{m}_k\}}} \sum_{k=1}^{K} \mathbbm{1}_{\{\boldsymbol{m}_k\}} \text{BCE}(\boldsymbol{m}_k, \boldsymbol{m}^*_k),
	\end{equation}
	\begin{equation}
	L_{\text{mask\_score}} = \frac{1}{\sum_{k=1}^{K} \mathbbm{1}_{\{\boldsymbol{r}_k\}}} \sum_{k=1}^{K} \mathbbm{1}_{\{\boldsymbol{r}_k\}} \Vert \boldsymbol{r}_k - \boldsymbol{r}_k^* \Vert_2.
	\end{equation}
	Here, ${c}^*$, $\boldsymbol{m}^*$, $\boldsymbol{r}^*$ are the classification, segmentation, and mask scoring targets, respectively. $K$ is the total number of proposals and $\mathbbm{1}_{\{.\}}$ indicates whether the proposal is a positive sample. 
	
	\subsection{Multi-task Learning}
	The whole network can be trained in an end-to-end manner using a multi-task loss.
	\begin{equation}
	L = L_\text{semantic} + L_\text{offset} + L_\text{class} + L_\text{mask} + L_\text{mask\_score},
	\end{equation}
	where $L_\text{semantic}$ and $L_\text{offset}$ are the semantic and offset losses defined at subsection \Cref{ssec:point_wise_net} while $L_\text{class}$, $L_\text{mask}$ and $L_\text{mask\_score}$ are the classification, segmentation and mask score losses defined at \Cref{ssec:top_down_refinement}.
	
	
	\begin{table*}[]
		\small
		\centering
		\setlength{\tabcolsep}{2.8pt}
		\begin{tabular}{@{}l|c|cccccccccccccccccc@{}} \toprule
			Method         & AP$_{50}$ & \rotatebox[origin=c]{90}{bathtub} & \rotatebox[origin=c]{90}{bed } & \rotatebox[origin=c]{90}{bookshe.} & \rotatebox[origin=c]{90}{cabinet} & \rotatebox[origin=c]{90}{chair} & \rotatebox[origin=c]{90}{counter} & \rotatebox[origin=c]{90}{curtain} & \rotatebox[origin=c]{90}{desk} & \rotatebox[origin=c]{90}{door} & \rotatebox[origin=c]{90}{other} & \rotatebox[origin=c]{90}{picture} & \rotatebox[origin=c]{90}{fridge} & \rotatebox[origin=c]{90}{s. curtain} & \rotatebox[origin=c]{90}{sink} & \rotatebox[origin=c]{90}{sofa} & \rotatebox[origin=c]{90}{table} & \rotatebox[origin=c]{90}{toilet} & \rotatebox[origin=c]{90}{window} \\ \midrule
			SGPN \cite{wang2018sgpn}          & 14.3          & 20.8           & 39.0          & 16.9          & 6.5           & 27.5          & 2.9           & 6.9           & 0.0           & 8.7           & 4.3           & 1.4           & 2.7           & 0.0            & 11.2          & 35.1          & 16.8          & 43.8           & 13.8          \\
			GSPN \cite{yi2019gspn}          & 30.6          & 50.0           & 40.5          & 31.1          & 34.8          & 58.9          & 5.4           & 6.8           & 12.6          & 28.3          & 29.0          & 2.8           & 21.9          & 21.4           & 33.1          & 39.6          & 27.5          & 82.1           & 24.5          \\
			3D-SIS \cite{hou20193d}        & 38.2          & \textbf{100.0} & 43.2          & 24.5          & 19.0          & 57.7          & 1.3           & 26.3          & 3.3           & 32.0          & 24.0          & 7.5           & 42.2          & 85.7           & 11.7          & 69.9          & 27.1          & 88.3           & 23.5          \\
			MASC \cite{liu2019masc}          & 44.7          & 52.8           & 55.5          & 38.1          & 38.2          & 63.3          & 0.2           & 50.9          & 26.0          & 36.1          & 43.2          & 32.7          & 45.1          & 57.1           & 36.7          & 63.9          & 38.6          & 98.0           & 27.6          \\
			PanopticFusion \cite{narita2019panopticfusion} & 47.8          & 66.7           & 71.2          & 59.5          & 25.9          & 55.0          & 0.0           & 61.3          & 17.5          & 25.0          & 43.4          & 43.7          & 41.1          & 85.7           & 48.5          & 59.1          & 26.7          & 94.4           & 35.9          \\
			3D-Bonet \cite{yang2019learning}      & 48.8          & \textbf{100.0} & 67.2          & 59.0          & 30.1          & 48.4          & 9.8           & 62.0          & 30.6          & 34.1          & 25.9          & 12.5          & 43.4          & 79.6           & 40.2          & 49.9          & 51.3          & 90.9           & 43.9          \\
			MTML \cite{lahoud20193d}          & 54.9          & \textbf{100.0} & 80.7          & 58.8          & 32.7          & 64.7          & 0.4           & 81.5          & 18.0          & 41.8          & 36.4          & 18.2          & 44.5          & \textbf{100.0} & 44.2          & 68.8          & 57.1          & \textbf{100.0} & 39.6          \\
			3D-MPA \cite{engelmann20203d}        & 61.1          & \textbf{100.0} & 83.3          & 76.5          & 52.6          & 75.6          & 13.6          & 58.8          & 47.0          & 43.8          & 43.2          & 35.8          & 65.0          & 85.7           & 42.9          & 76.5          & 55.7          & \textbf{100.0} & 43.0          \\
			Dyco3D \cite{he2021dyco3d}        & 64.1          & \textbf{100.0} & 84.1          & \textbf{89.3} & 53.1          & 80.2          & 11.5          & 58.8          & 44.8          & 43.8          & 53.7          & 43.0          & 55.0          & 85.7           & 53.4          & 76.4          & 65.7          & 98.7           & 56.8          \\
			PE \cite{zhang2019point}            & 64.5          & \textbf{100.0} & 77.3          & 79.8          & 53.8          & 78.6          & 8.8           & 79.9          & 35.0          & 43.5          & 54.7          & 54.5          & 64.6          & 93.3           & 56.2          & 76.1          & 55.6          & 99.7           & 50.1          \\
			PointGroup \cite{jiang2020pointgroup}    & 63.6          & \textbf{100.0} & 76.5          & 62.4          & 50.5          & 79.7          & 11.6          & 69.6          & 38.4          & 44.1          & 55.9          & 47.6          & 59.6          & \textbf{100.0} & 66.6          & 75.6          & 55.6          & 99.7           & 51.3          \\
			GICN \cite{liu2020learning}          & 63.8          & \textbf{100.0} & \textbf{89.5} & 80.0          & 48.0          & 67.6          & 14.4          & 73.7          & 35.4          & 44.7          & 40.0          & 36.5          & 70.0          & \textbf{100.0} & 56.9          & 83.6          & 59.9          & \textbf{100.0} & 47.3          \\
			OccuSeg \cite{han2020occuseg}       & 67.2          & \textbf{100.0} & 75.8          & 68.2          & 57.6          & 84.2          & \textbf{47.7} & 50.4          & 52.4          & 56.7          & 58.5          & 45.1          & 55.7          & \textbf{100.0} & 75.1          & 79.7          & 56.3          & \textbf{100.0} & 46.7          \\
			SSTNet \cite{liang2021instance}        & 69.8          & \textbf{100.0} & 69.7          & 88.8          & 55.6          & 80.3          & 38.7          & 62.6          & 41.7          & 55.6          & 58.5          & \textbf{70.2} & 60.0          & \textbf{100.0} & \textbf{82.4} & 72.0          & 69.2          & \textbf{100.0} & 50.9          \\
			HAIS \cite{chen2021hierarchical}          & 69.9          & \textbf{100.0} & 84.9          & 82.0          & 67.5          & 80.8          & 27.9          & 75.7          & 46.5          & 51.7          & 59.6          & 55.9          & 60.0          & \textbf{100.0} & 65.4          & 76.7          & 67.6          & 99.4           & 56.0          \\
			\textbf{SoftGroup (Ours)}     & \textbf{76.1} & \textbf{100.0} & 80.8          & 84.5          & \textbf{71.6} & \textbf{86.2} & 24.3          & \textbf{82.4} & \textbf{65.5} & \textbf{62.0} & \textbf{73.4} & 69.9          & \textbf{79.1} & 98.1           & 71.6          & \textbf{84.4} & \textbf{76.9} & \textbf{100.0} & \textbf{59.4} \\ \bottomrule
		\end{tabular}
		\caption{3D instance segmentation results on ScanNet v2 hidden test set in terms of AP$_{50}$ scores. The proposed SoftGroup achieves the highest average AP$_{50}$, outperforming the previous strongest method by a significant margin. Reported results are from the ScanNet benchmark on 13/11/2021.}
		\label{tab:scannet_benchmark}
	\end{table*}
	
	\begin{figure*}
		\centering
		\begin{subfigure}{0.195\textwidth}
			\includegraphics[width=\textwidth]{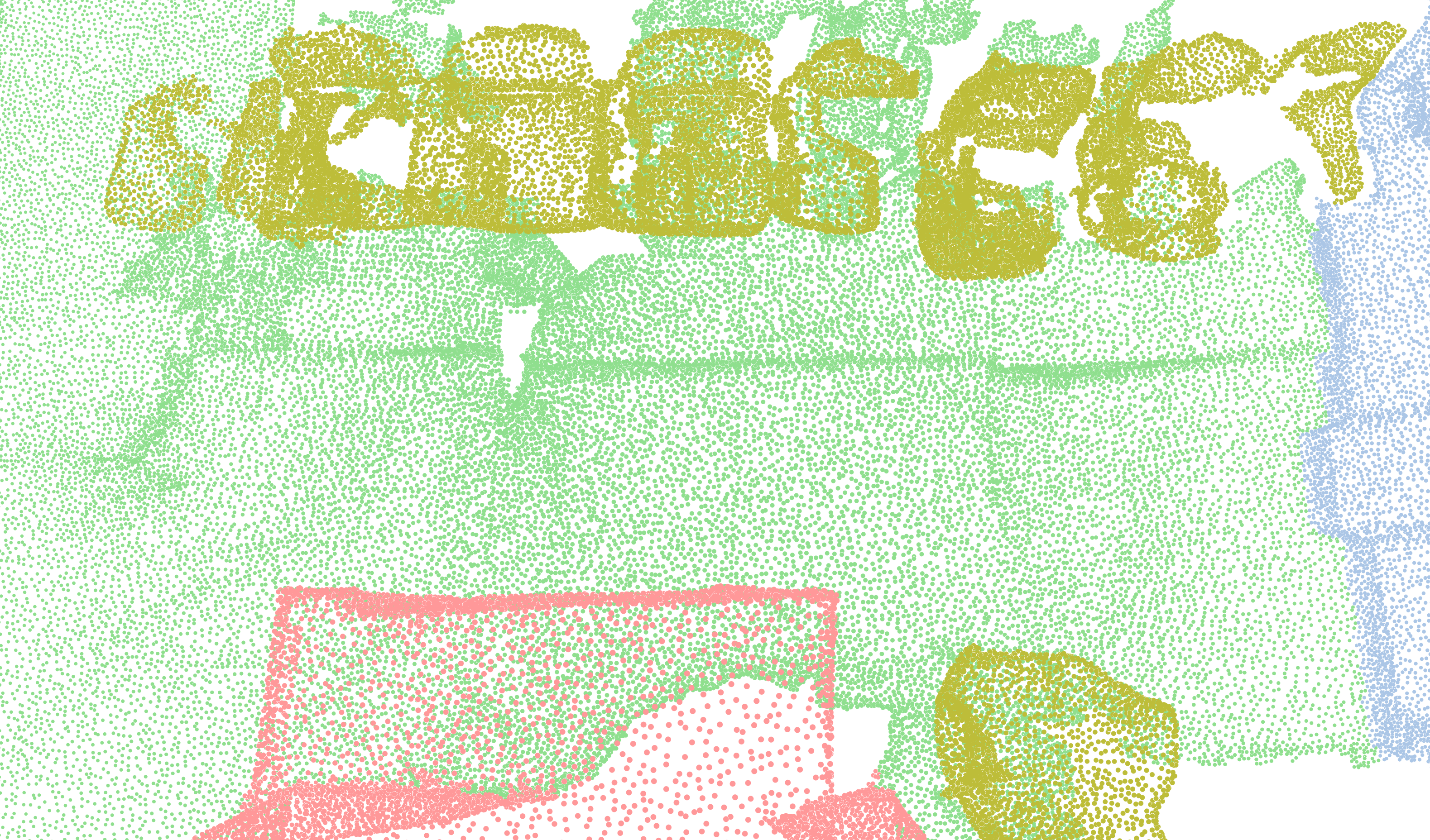}
		\end{subfigure}
		\begin{subfigure}{0.195\textwidth}
			\includegraphics[width=\textwidth]{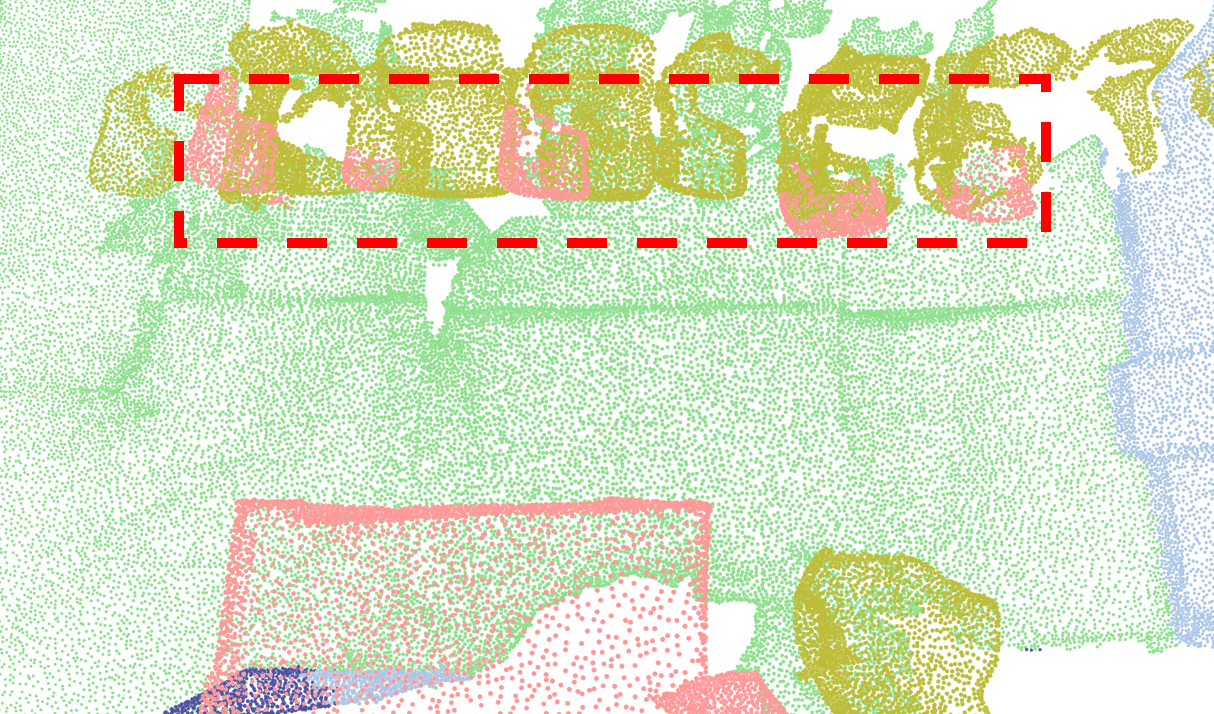}
		\end{subfigure}
		\begin{subfigure}{0.195\textwidth}
			\includegraphics[width=\textwidth]{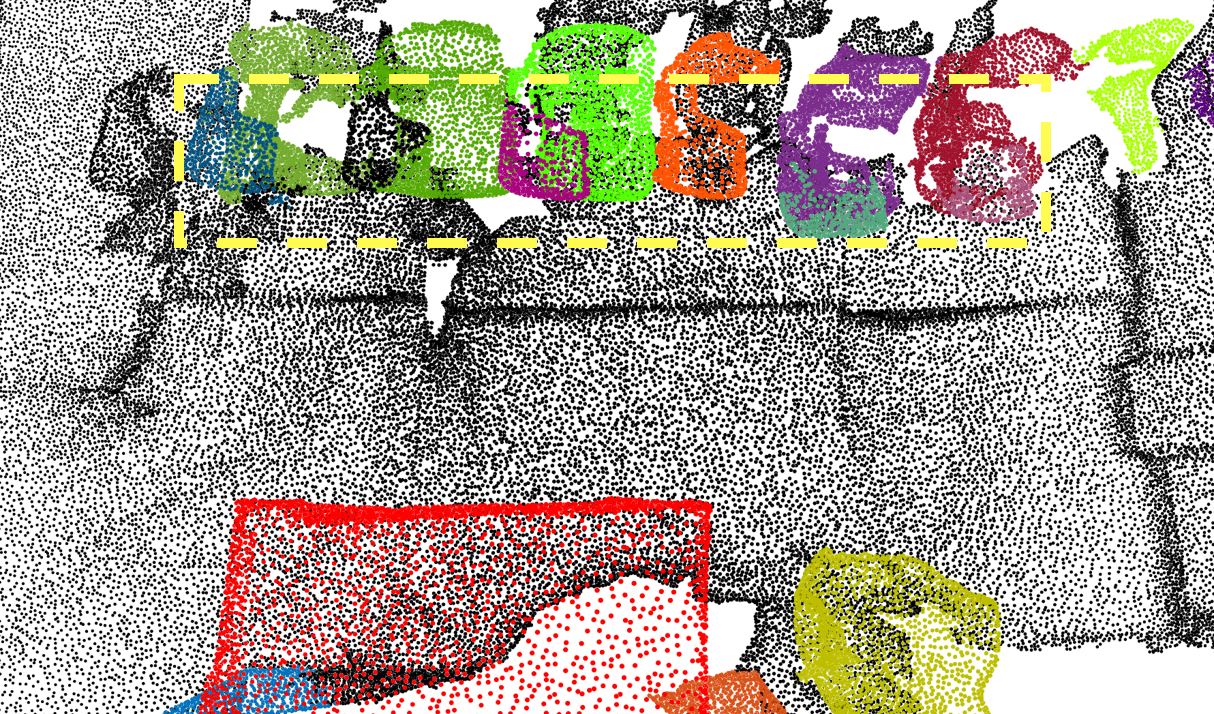}
		\end{subfigure}
		\begin{subfigure}{0.195\textwidth}
			\includegraphics[width=\textwidth]{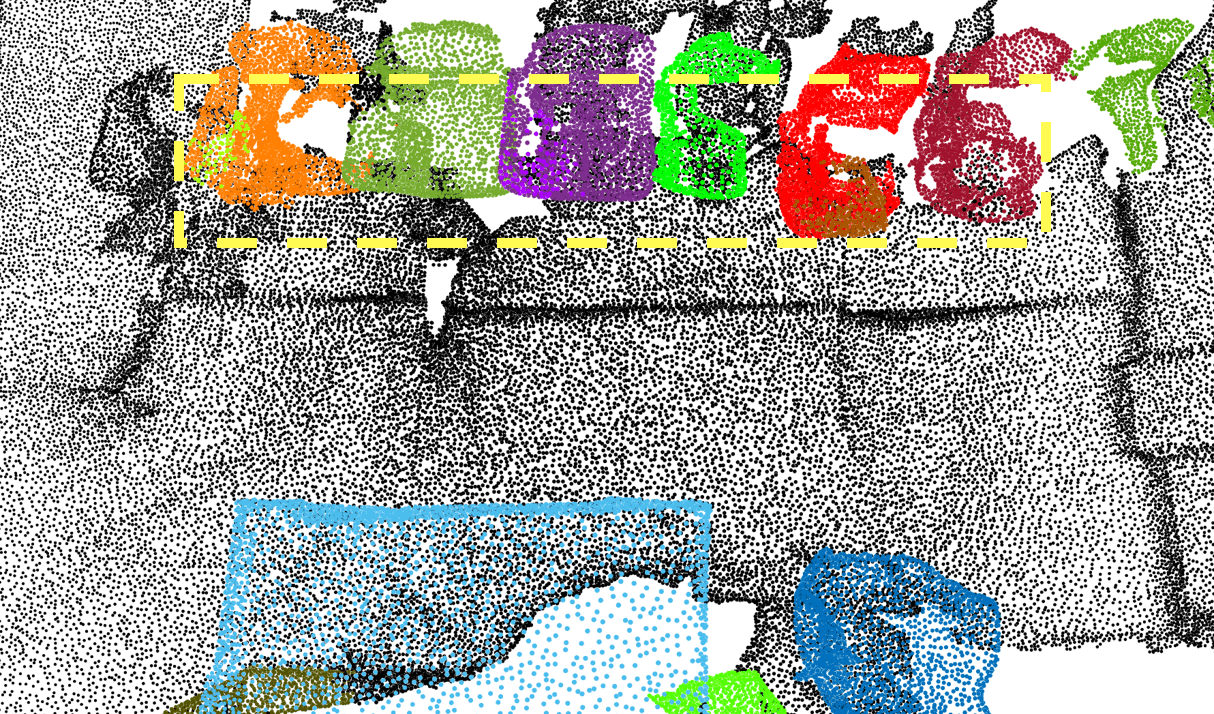}
		\end{subfigure}
		\begin{subfigure}{0.195\textwidth}
			\includegraphics[width=\textwidth]{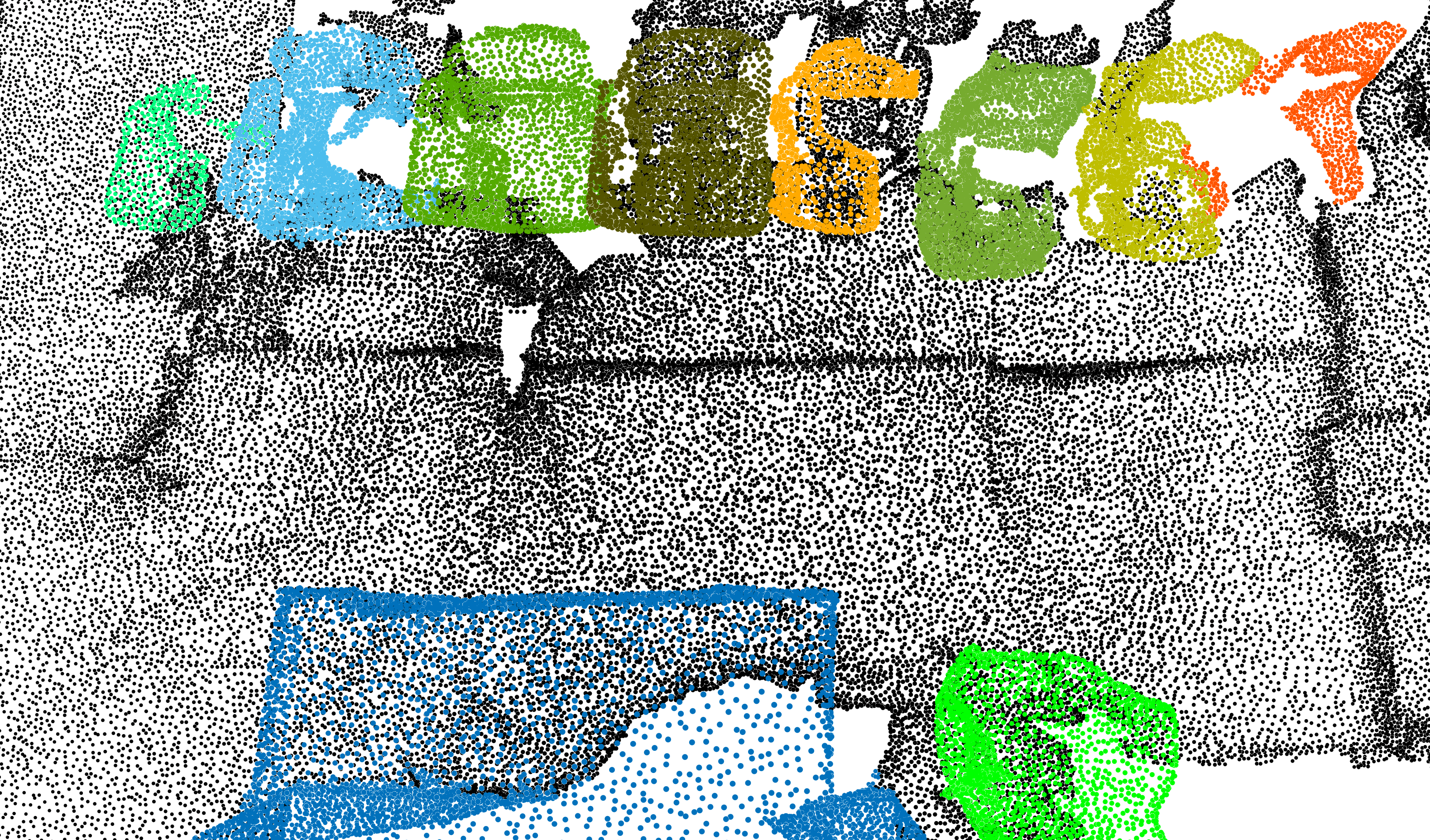}
		\end{subfigure}
		
		\begin{subfigure}{0.195\textwidth}
			\includegraphics[width=\textwidth]{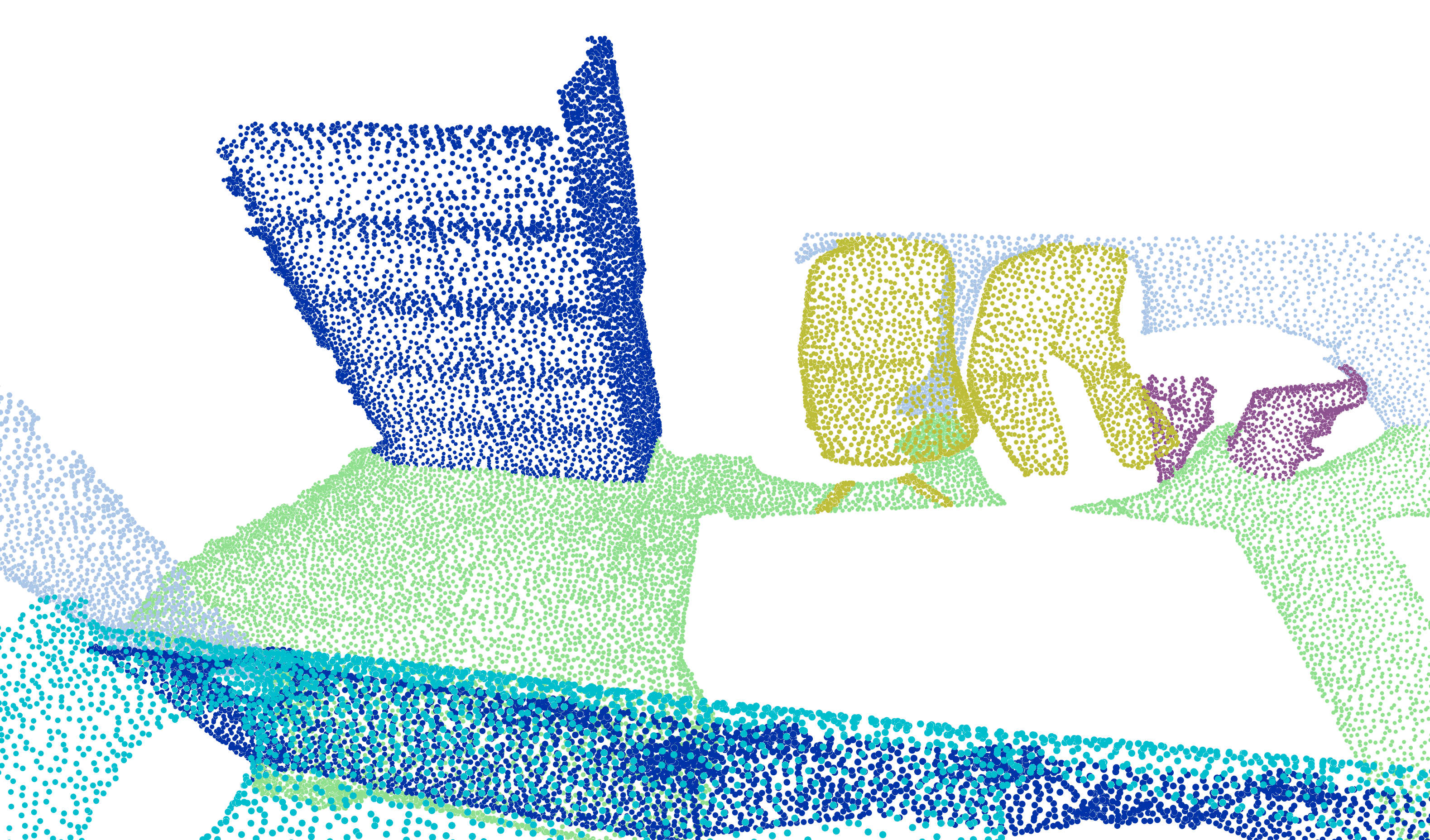}
		\end{subfigure}
		\begin{subfigure}{0.195\textwidth}
			\includegraphics[width=\textwidth]{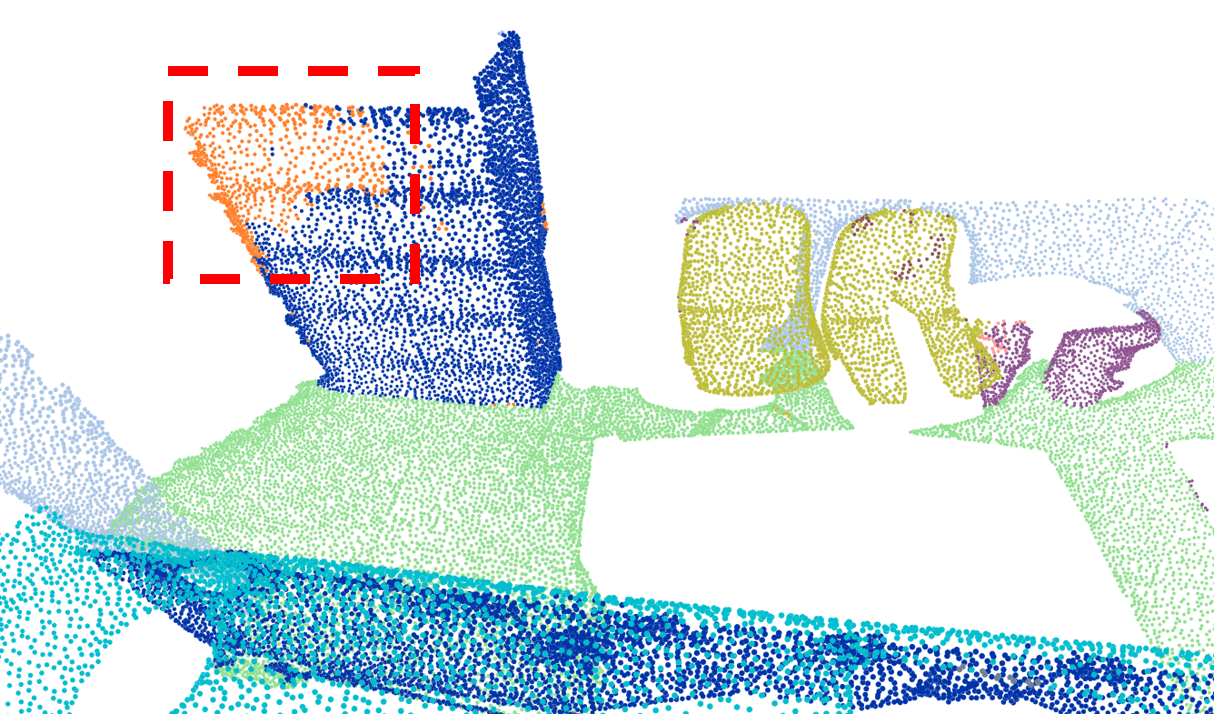}
		\end{subfigure}
		\begin{subfigure}{0.195\textwidth}
			\includegraphics[width=\textwidth]{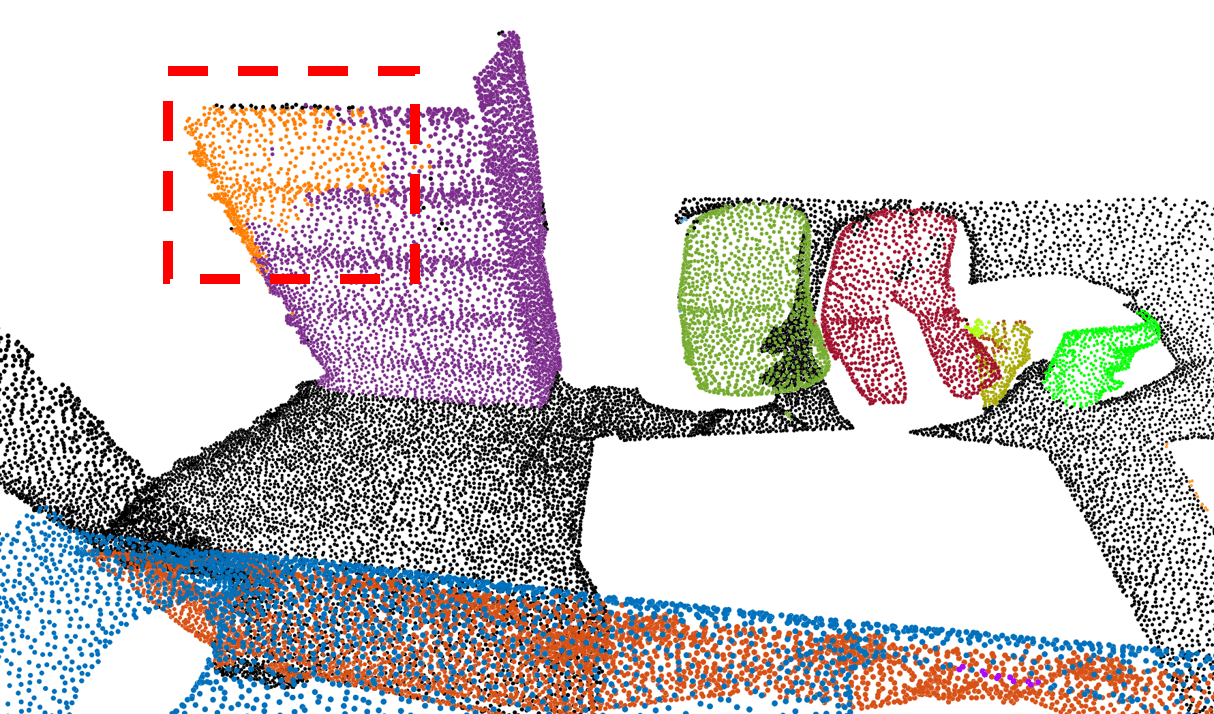}
		\end{subfigure}
		\begin{subfigure}{0.195\textwidth}
			\includegraphics[width=\textwidth]{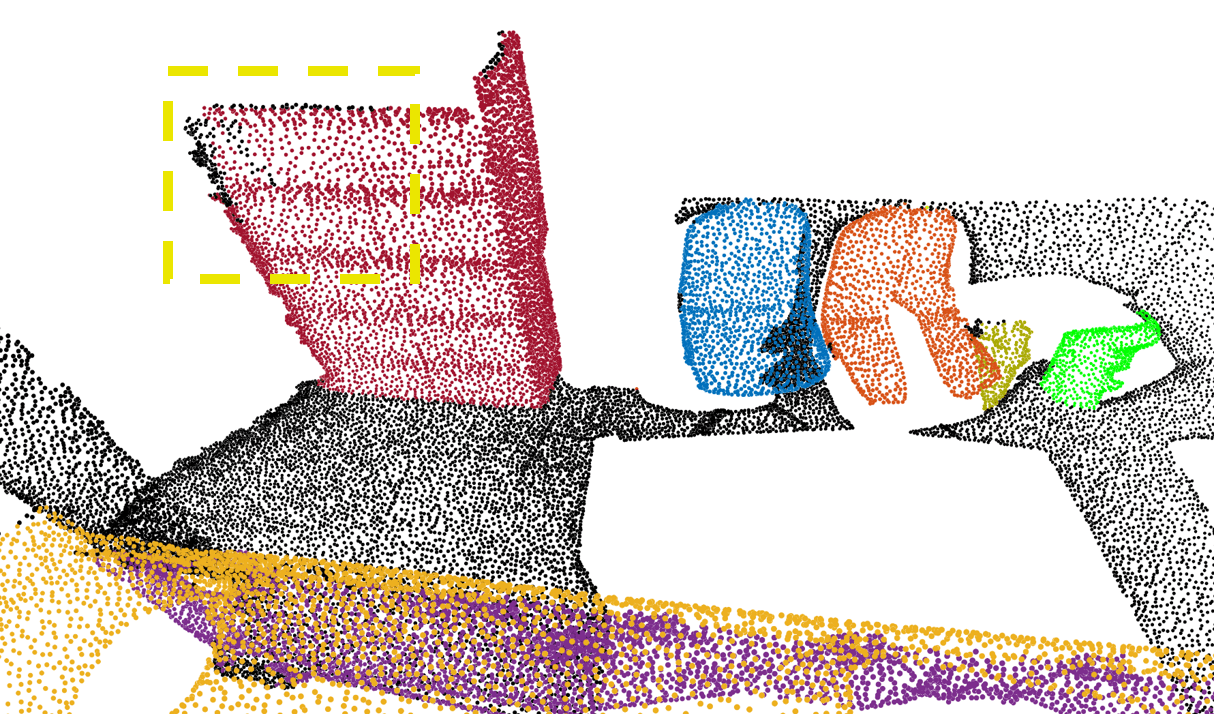}
		\end{subfigure}
		\begin{subfigure}{0.195\textwidth}
			\includegraphics[width=\textwidth]{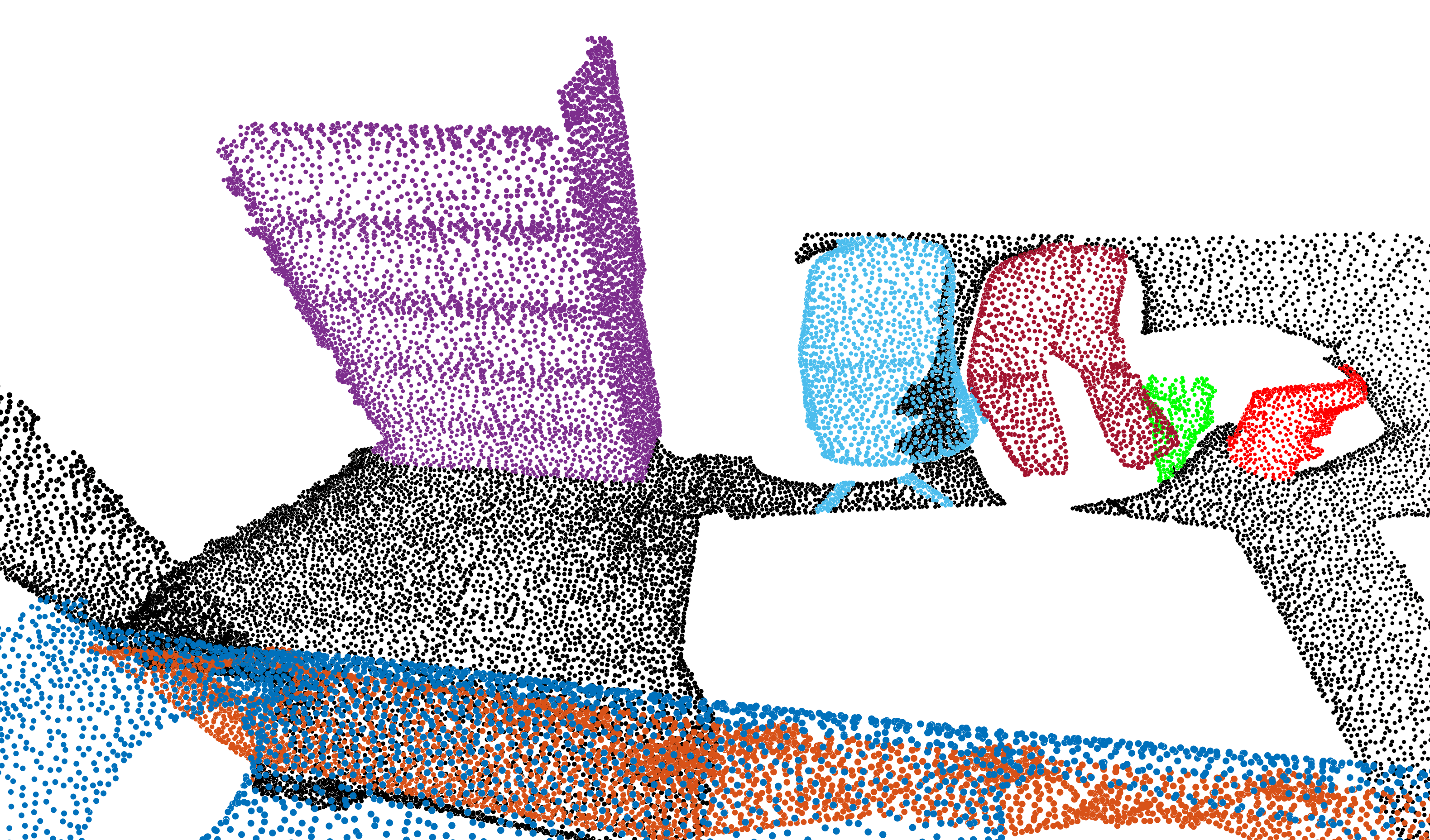}
		\end{subfigure}
		
		\begin{subfigure}{0.195\textwidth}
			\includegraphics[width=\textwidth]{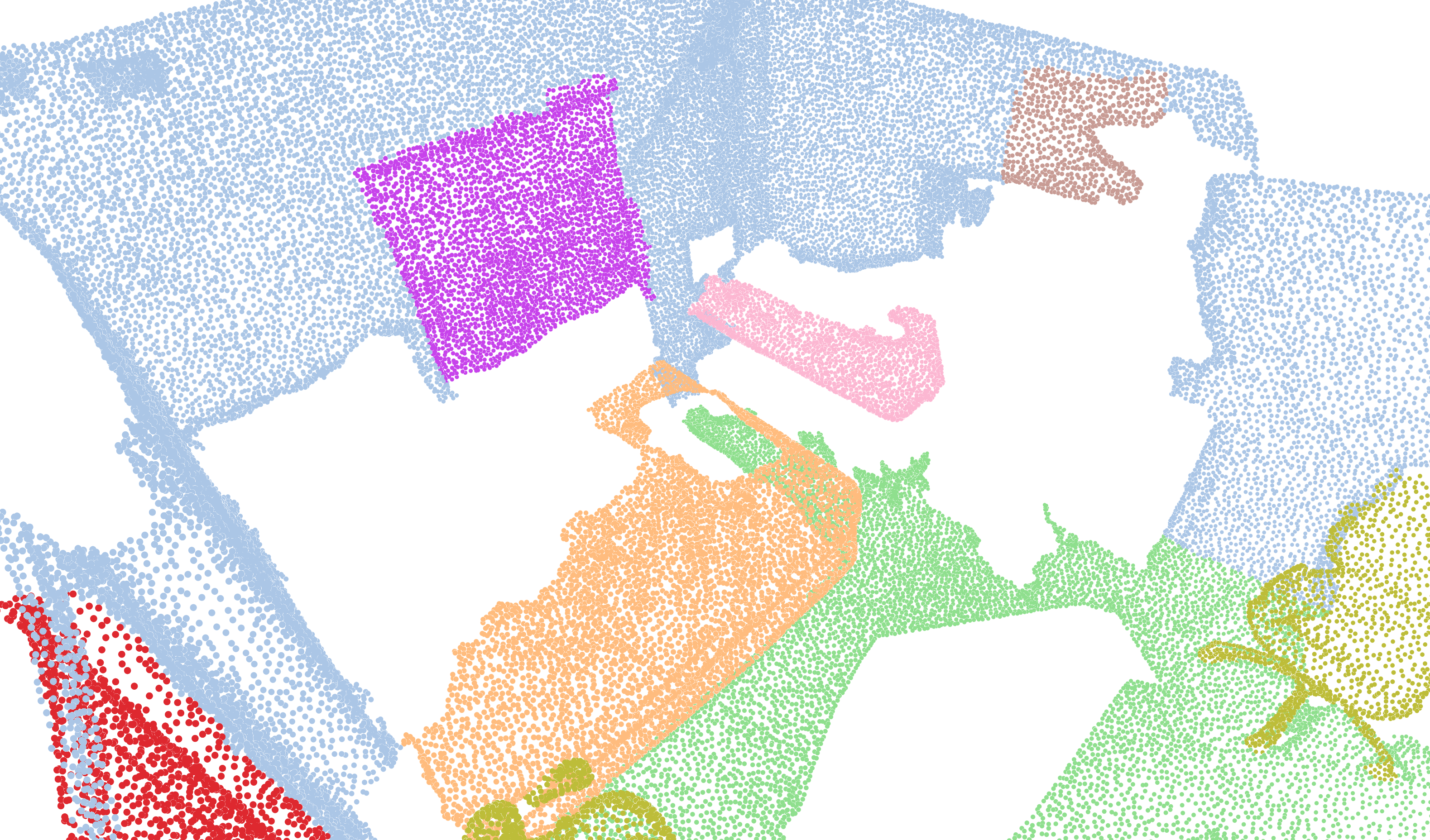}
			\caption*{Semantic GT}
		\end{subfigure}
		\begin{subfigure}{0.195\textwidth}
			\includegraphics[width=\textwidth]{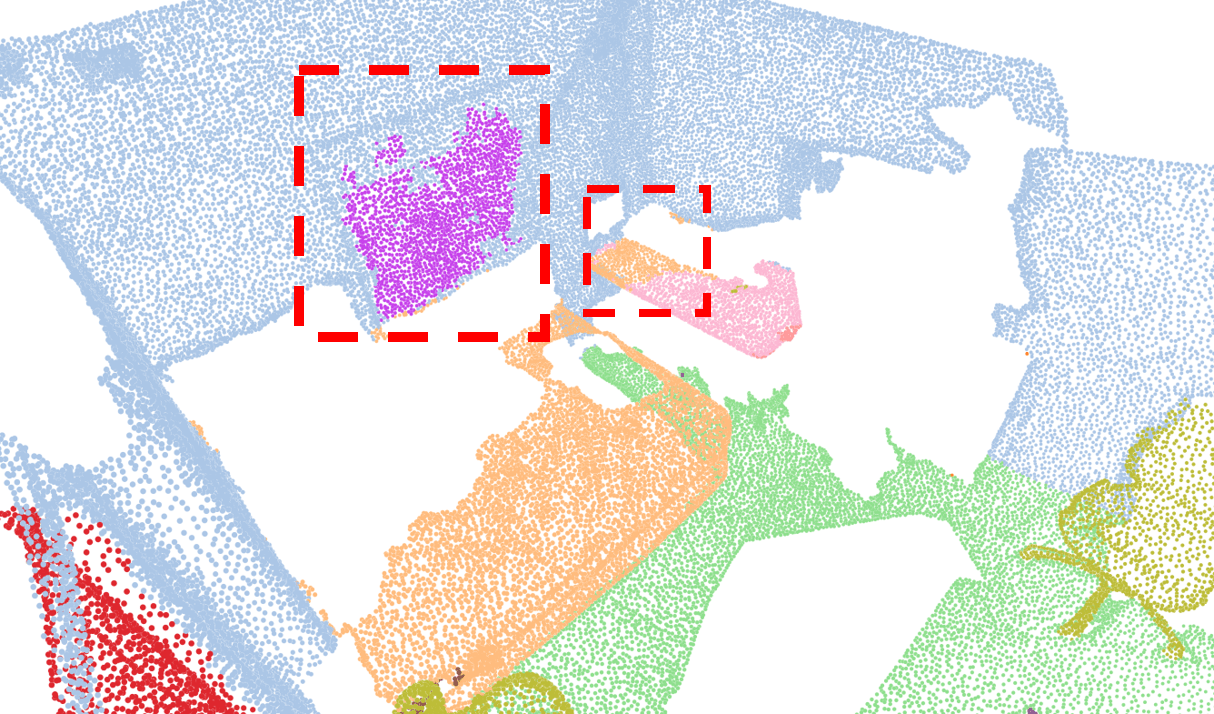}
			\caption*{Semantic pred}
		\end{subfigure}
		\begin{subfigure}{0.195\textwidth}
			\includegraphics[width=\textwidth]{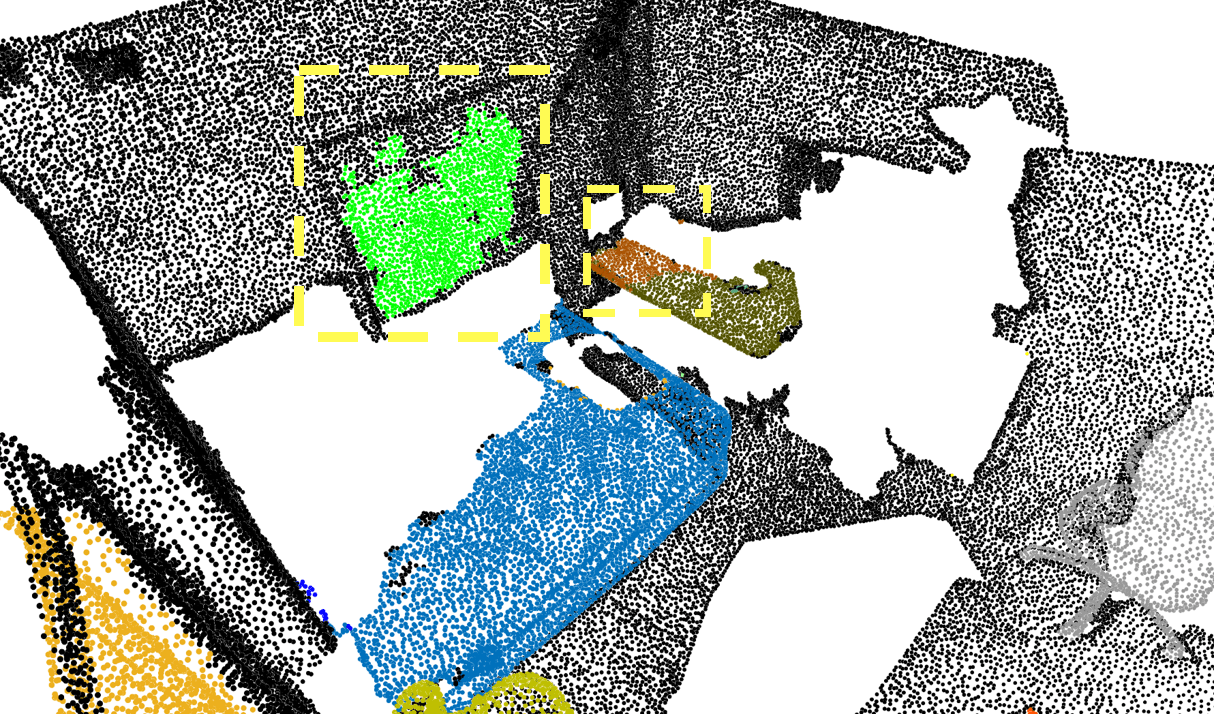}
			\caption*{Instance pred w/o SoftGroup}
		\end{subfigure}
		\begin{subfigure}{0.195\textwidth}
			\includegraphics[width=\textwidth]{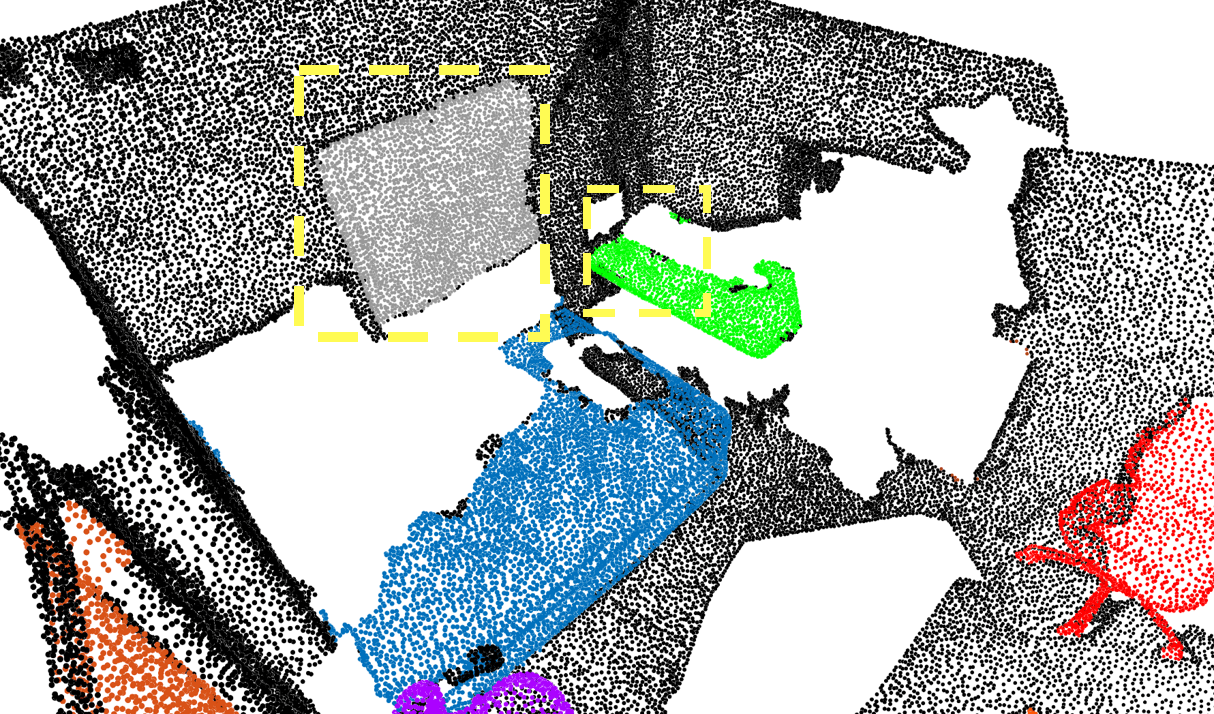}
			\caption*{Instance pred w/ SoftGroup}
		\end{subfigure}
		\begin{subfigure}{0.195\textwidth}
			\includegraphics[width=\textwidth]{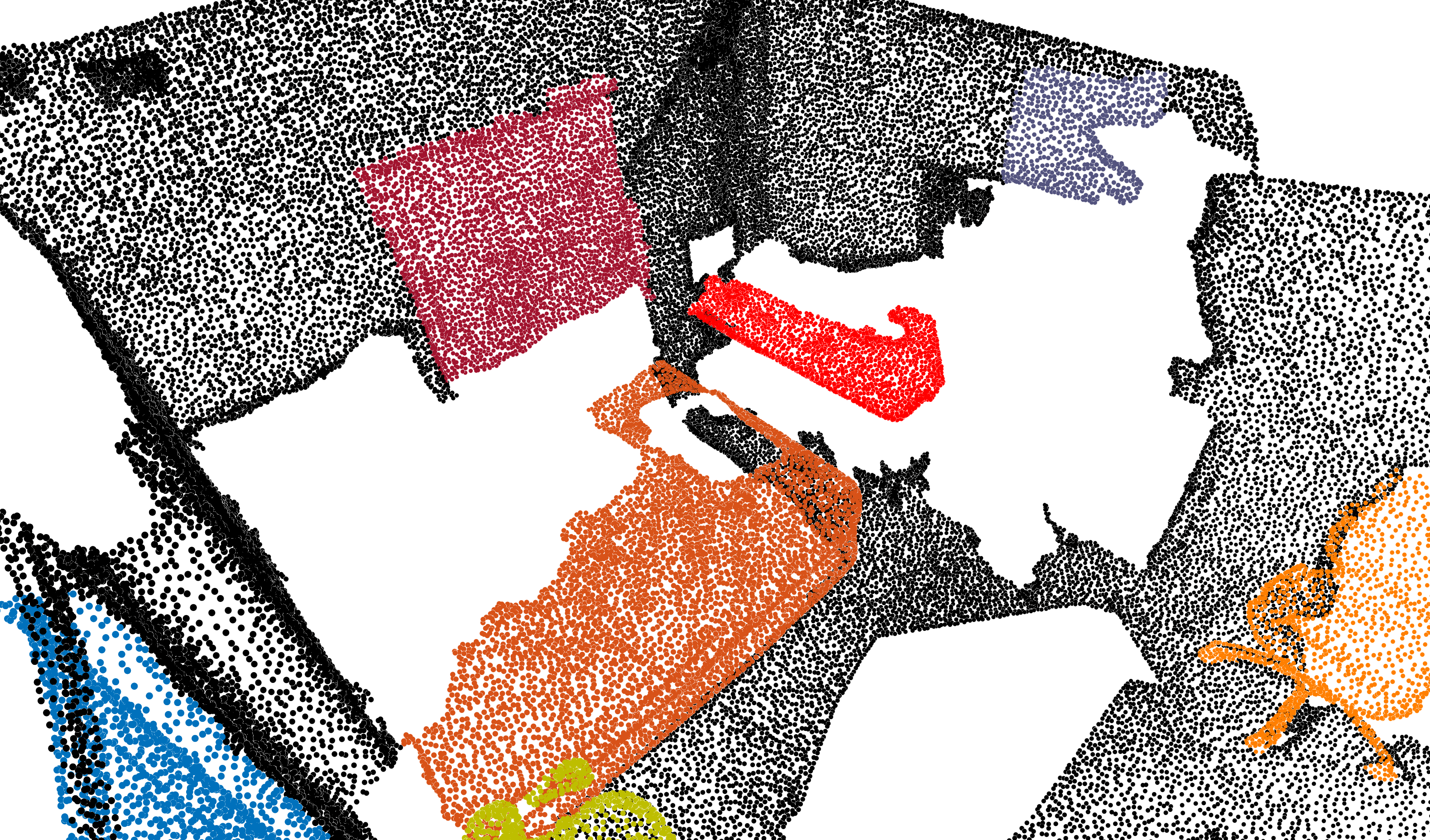}
			\caption*{Instance GT}
		\end{subfigure}
		\caption{Qualitative results on ScanNet v2 validation set. Instance prediction without SoftGroup output low-quality instance mask at the region of wrong semantic prediction (highlighted by dashed boxes). The prediction with SoftGroup shows more accurate instance masks at these regions. \textbf{Zoom in for best view}.}
		\label{fig:quanlitative_res}
	\end{figure*}
	
	\begin{table}[]
		\small
		\setlength{\tabcolsep}{2.8pt}
		\begin{tabular}{@{}lcccccc@{}} \toprule
			Method     & AP   & AP$_{50}$ & mCov & mWCov & mPrec$_{50}$ & mRec$_{50}$ \\ \midrule
			SGPN$^{\dagger}$ \cite{wang2018sgpn}       &   -   &  -    & 32.7 & 35.5  & 36.0    & 28.7   \\
			ASIS$^{\dagger}$ \cite{wang2019associatively}       &  -    &  -    & 44.6 & 47.8  & 55.3    & 42.4   \\
			PointGroup$^{\dagger}$ \cite{jiang2020pointgroup} &  -    & 57.8 &   -   &   -    & 61.9    & 62.1   \\
			SSTNet$^{\dagger}$ \cite{liang2021instance}     & 42.7 & 59.3 &   -   &  -     & 65.5    & 64.2   \\
			HAIS$^{\dagger}$ \cite{chen2021hierarchical}       &   -   &   -   & 64.3 & 66.0  & 71.1    & 65.0   \\
			\textbf{SoftGroup}$^{\dagger}$  & \textbf{51.6} & \textbf{66.1} & \textbf{66.1} & \textbf{68.0}  & \textbf{73.6}    & \textbf{66.6}   \\ \midrule
			SGPN$^{\ddagger}$ \cite{wang2018sgpn}       &  -    &    -  & 37.9 & 40.8  & 38.2    & 31.2   \\
			PartNet$^{\ddagger}$ \cite{mo2019partnet}    &   -   &    -  &   -   &     -  & 56.4    & 43.4   \\
			ASIS$^{\ddagger}$ \cite{wang2019associatively}       &   -   &   -   & 51.2 & 55.1  & 63.6    & 47.5   \\
			3D-BoNet$^{\ddagger}$ \cite{yang2019learning}   &  -    &   -   &   -   &     -  & 65.6    & 47.7   \\
			OccuSeg$^{\ddagger}$ \cite{han2020occuseg}    &   -   &  -    &    -  &    -   & 72.8    & 60.3   \\
			GICN$^{\ddagger}$ \cite{liu2020learning}       &   -   &   -   &    -  &    -   & 68.5    & 50.8   \\
			PointGroup$^{\ddagger}$ \cite{jiang2020pointgroup} &   -   & 64.0 & -     &  -     & 69.6    & 69.2   \\
			SSTNet$^{\ddagger}$  \cite{liang2021instance}     & 54.1 & 67.8 & -     &   -    & 73.5    & \textbf{73.4}   \\
			HAIS$^{\ddagger}$  \cite{chen2021hierarchical}      &   -   &    -  & 67.0 & 70.4  & 73.2    & 69.4   \\
			\textbf{SoftGroup}$^{\ddagger}$  & \textbf{54.4} & \textbf{68.9} & \textbf{69.3} & \textbf{71.7}  & \textbf{75.3}    & 69.8  \\ \bottomrule
		\end{tabular}
		\caption{3D instance segmentation results on S3DIS dataset. Methods marked with ${\dagger}$ are evaluated on Area 5, and methods marked with  ${\ddagger}$ are evaluated on 6-fold cross validation.}
		\label{tab:s3dis_benchmark}
	\end{table}
	
	\begin{table}[]
		\setlength{\tabcolsep}{5pt}
		\centering
		\begin{tabular}{lcccc} \toprule
			Method     & AP$_{50}$ & AP$_{25}$ & Box AP$_{50}$ & Box AP$_{25}$ \\ \midrule
			F-PointNet \cite{qi2018frustum} & -      & -      & 10.8    & 19.8    \\
			GSPN \cite{yi2019gspn}       & 37.8   & 53.4   & 17.7    & 30.6    \\
			3D-SIS \cite{hou20193d}    & 18.7   & 35.7   & 22.5    & 40.2    \\
			VoteNet \cite{qi2019deep}   & -      & -      & 33.5    & 58.6    \\
			3D-MPA \cite{engelmann20203d}    & 51.9   & 72.4   & 49.2    & 64.2    \\
			PointGroup \cite{jiang2020pointgroup} & 51.7   & 71.3   & 48.9    & 61.5    \\
			SSTNet \cite{liang2021instance}    & 64.3   & 74.0     & 52.7    & 62.5    \\
			HAIS \cite{chen2021hierarchical}      & 64.4   & 75.6   & 53.1    & 64.3    \\
			\textbf{SoftGroup} & \textbf{67.6}   & \textbf{78.9}   & \textbf{59.4}    & \textbf{71.6}    \\ \bottomrule 
		\end{tabular}
		\caption{Instance segmentation and object detection results on ScanNet v2 validation set. Our method achieves better results on both mask and box AP.}
		\label{tab:segmentation_detection}
	\end{table}

	\section{Experiments}
	\subsection{Experimental Settings}
	\paragraph{Datasets.}
	The experiments are conducted on standard benchmarked ScanNet v2 \cite{dai2017scannet} and S3DIS \cite{armeni20163d} dataset. The ScanNet dataset contains 1613 scans which is divided into training, validation, and testing sets of 1201, 312, 100 scans, respectively. Instance segmentation is evaluated on 18 object classes. Following existing methods, the benchmarked results are reported on the hidden test split. The ablation study is conducted on the validation set. 
	
	The S3DIS dataset contains 3D scans of 6 areas with 271 scenes in total. The dataset consists of 13 classes for instance segmentation evaluation. Following existing methods, two settings are used to evaluate the instance segmentation results: testing on Area 5 and 6-fold cross-validation.
	
	\paragraph{Evaluation Metrics.} The evaluation metric is the standard average precision. Here, AP$_{50}$ and AP$_{25}$ denote the scores with IoU thresholds of 50\% and 25\%, respectively. Likewise, AP denotes the averaged scores with IoU threshold from 50\% to 95\% with a step size of 5\%. Additionally, the S3DIS is also evaluated using mean coverage (mCov), mean weighed coverage (mWCov), mean precision (mPrec), and mean recall (mRec). 
	
	\paragraph{Implementation Details.}
	The implementation details follow those of existing methods \cite{jiang2020pointgroup,chen2021hierarchical}. The model is implemented using PyTorch deep learning framework \cite{paszke2017automatic} and trained on 120k iterations with Adam optimizer \cite{kingma2014adam}. The batch size is set to 4. The learning rate is initialized to 0.001 and scheduled by a cosine annealing \cite{loshchilov2016sgdr}. The voxel size and grouping bandwidth $b$ are set to 0.02m and 0.04m, respectively. The score threshold for soft grouping $\tau$ is set to 0.2. At training time, the scenes are randomly cropped at a maximum number of points of 250k. At inference, the whole scene is fed into the network without cropping. For the S3DIS with high point density, scenes are randomly downsampled at a ratio of 1/4 before cropping. At inference, the scene is divided into four parts before feeding into the model, and then the outputs from the four parts are merged to get the final results. 
	
	
	We note that the source code and trained models for existing high-performing methods are publicly available on ScanNet v2 only. In this work, the source code and trained models on both ScanNet v2 and S3DIS will be released to support result reproducibility.

	\begin{table}[]
		\small
		\centering
		\setlength{\tabcolsep}{4pt}
		\begin{tabular}{llc@{}} \toprule
			Method     & Component time (ms)                    & Total   (ms) \\ \midrule
			& Backbone   (GPU): 2080                   &              \\
			SGPN \cite{wang2018sgpn}      & Group   merging (CPU): 149000            & 158439       \\
			& Block   merging (CPU): 7119              &              \\  \midrule
			& Backbone   (GPU): 2083                   &              \\
			ASIS \cite{wang2019associatively}      & Mean   shift (CPU): 172711               & 181913       \\
			& Block   merging (CPU): 7119              &              \\  \midrule
			& Backbone   (GPU): 1612                   &              \\
			GSPN \cite{yi2019gspn}      & Point   sampling (GPU): 9559             & 12702        \\
			& Neighbour   search (CPU): 1500           &              \\  \midrule
			& Backbone   (GPU): 2083                   &              \\
			3D-BoNet \cite{yang2019learning}   & SCN   (GPU): 667                         & 9202         \\
			& Block   merging (CPU): 7119              &              \\  \midrule
			& Backbone   (GPU): 1497                   &              \\
			GICN \cite{liu2020learning}      & SCN   (GPU): 667                         & 8615         \\
			& Block   merging(CPU): 7119               &              \\  \midrule
			& Backbone   GPU): 189                     &              \\
			OccuSeg \cite{han2020occuseg}   & Supervoxel   (CPU): 1202                 & 1904         \\
			& Clustering   (GPU+CPU): 513              &              \\  \midrule
			& Backbone   (GPU): 128                    &              \\
			PointGroup \cite{jiang2020pointgroup} & Clustering   (GPU+CPU):221               & 452          \\
			& ScoreNet   (GPU): 103                    &              \\  \midrule
			& Backbone (GPU) 125                       &              \\
			SSTNet \cite{liang2021instance}    & Tree network (GPU+CPU): 229              & 428          \\
			& ScoreNet (GPU): 74                       &              \\  \midrule
			& Pointwise   prediction (GPU): 154       &              \\
			HAIS \cite{chen2021hierarchical}      & Hier. aggr.  (GPU+CPU): 118   & \textbf{339}          \\
			& Intra-inst. prediction (GPU): 67      &              \\  \midrule
			\multirow{3}{*}{\begin{tabular}[l]{@{}l@{}}\textbf{SoftGroup}\\\textbf{(Ours)}\end{tabular}} & Pointwise   prediction (GPU): 152       &              \\
			& Soft grouping (GPU+CPU): 123             & 345          \\
			& Top-down refinement (GPU): 70                &       \\ \bottomrule
		\end{tabular}
		\caption{Inference time per scan of different methods on ScanNet v2 validation set. For a fair comparison, the runtime is measured on the same Titan X GPU model.}
		\label{tab:runtime}
	\end{table}

	\subsection{Benchmarking Results}
	\paragraph{ScanNet v2.}
	Table \ref{tab:scannet_benchmark} shows the results of SoftGroup and recent state-of-the-art methods on the hidden test set of ScanNet v2 benchmark. We submit our model and report the results from the server.  The proposed SoftGroup achieves the highest average AP$_{50}$ of 76.1\%, surpassing the previous strongest methods a significant margin of 6.2\%. Regarding class-wise scores, our method achieves the best performance in 12 out of 18 classes.
	
	\paragraph{S3DIS.}
	Table \ref{tab:s3dis_benchmark} summaries the results on Area 5 and 6-fold cross-validation of S3DIS dataset. On both Area 5 and cross-validation evaluations, the proposed SoftGroup achieves higher overall performance compared to existing method. Notably, on Area 5 evaluation, SoftGroup achieves AP/AP$_{50}$ of 51.6/66.1(\%), which is 8.9/6.8(\%) improvement compared to the second-best. The state-of-the-art performance on both ScanNet v2 and S3DIS datasets shows the generalization advantage of our method.

	\paragraph{Segmentation and Detection Results.}
	We further report the instance segmentation and object detection results on ScanNet v2 validation set. To obtain object detection results, we follow the approach in \cite{engelmann20203d} to extract a tight axis-aligned bounding box from the predicted point mask. Table \ref{tab:segmentation_detection} reports the instance segmentation and object detection results. Our method achieves significant improvement compared to the second-best by 3.2, 3.3, 6.3, and 7.3(\%) of AP$_{50}$, AP$_{25}$, box AP$_{50}$, and box AP$_{25}$, respectively.
	
	\paragraph{Runtime Analysis.}
	Table \ref{tab:runtime} report the runtime per scan of different methods on ScanNet v2 validation set. For a fair comparison, the reported runtime is measured on the same Titan X GPU model. The inference time of our method is 345ms per scan, which is extra 6ms over the fastest model. Regarding our component-time, the point-wise prediction network, soft grouping algorithm, and top-down refinement latencies are 152ms, 132ms, and 70ms, respectively. The results show that our method achieves high accuracy while remaining computationally efficient. 
	
	\subsection{Qualitative Analysis}
	Figure \ref{fig:quanlitative_res} shows the visualization examples from ScanNet v2 dataset. Without SoftGroup, the semantic prediction errors are propagated to instance segmentation predictions (highlighted by dashed boxes). In contrast, SoftGroup effectively corrects the semantic prediction errors and thus generates more accurate instance masks.
	
	\subsection{Ablation Study}
	\paragraph{Component-wise Analysis.}
	We provide experimental results of SoftGroup when different components are omitted. The considered baseline is a model with hard grouping and the confidence scores of output instances are ranked by a ScoreNet branch \cite{jiang2020pointgroup,liang2021instance}. Table \ref{tab:component_analysis} shows the ablation results. The baseline achieves 39.5/61.1/75.5(\%) in terms of AP/AP$_{50}$/AP$_{25}$. Significant improvement is obtained by either applying soft grouping or top-down refinement. Combining these two components achieves the best overall performance AP/AP$_{50}$/AP$_{25}$ of 46.0/67.6/78.9(\%), which is significantly higher than the baseline by 6.5/6.5/3.4(\%).
	
	\begin{table}[]
		\small
		\centering
		\setlength{\tabcolsep}{2.3pt}
		\begin{tabular}{@{}cccccc@{}}
			\toprule
			Baseline & Soft grouping & Top-down refinement & AP   & AP$_{50}$ & AP$_{25}$ \\ \midrule
			\checkmark        &               &                 & 39.5 & 61.1 & 75.5 \\
			& \checkmark             &                 & 41.6 & 63.8 & \textbf{79.2} \\
			&               & \checkmark               & 44.3 & 65.4 & 78.1 \\
			& \checkmark             & \checkmark               & \textbf{46.0}   & \textbf{67.6} & 78.9 \\ \midrule
			\multicolumn{3}{l}{\textbf{Overall improvement}} & \textbf{+6.5} & \textbf{+6.5} & \textbf{+3.4} \\ \bottomrule
		\end{tabular}
		\caption{Component-wise analysis on ScanNet v2 validation set. Our model achieves significant improvement over the baseline. }
		\label{tab:component_analysis}
	\end{table}
	
	\begin{table}[]
		\centering
		\setlength{\tabcolsep}{12pt}
		\begin{tabular}{c|ccc}
			\toprule
			$\tau$  & AP            & AP$_{50}$ & AP$_{25}$        \\ \midrule
			None & 44.3          & 65.4          & 78.1          \\ \midrule
			0.01 & 40.1          & 58.5          & 69.2          \\
			0.1  & 45.3          & 66.5          & 78.5          \\
			0.2  & \textbf{46.0} & \textbf{67.6} & \textbf{78.9} \\
			0.3  & 45.2          & 66.8          & 78.5          \\
			0.4  & 44.7          & 46.1          & 78.3          \\
			0.5  & 43.9          & 64.8          & 77.7          \\ \bottomrule
		\end{tabular}
		\caption{Ablation experiments on varying score threshold $\tau$ for soft grouping. ``None" denotes the threshold is not used, and the hard semantic prediction is used for grouping instead.}
		\label{tab:score_thr}
	\end{table}
	
	\paragraph{Score Threshold for Soft Grouping.}
	Table \ref{tab:score_thr} shows the experimental results with varying score thresholds for soft grouping. The baseline is with $\tau$ being ``None", indicating the threshold is deactivated and the hard predicted label is used for grouping. The baseline achieves AP/AP$_{50}$/AP$_{25}$ of 44.3/65.4/78.1(\%). When $\tau$ is too high or too low the performance is even worse than the baseline. The best performance is obtained at $\tau$ of 0.2, which confirms our analysis at the \Cref{ssec:softgroup}, where the number of positive and negative samples are balanced.

	\begin{table}[]
		\centering
		\begin{tabular}{cccccc}
			\toprule
			Class & Mask& Mask score& AP          &AP$_{50}$ & AP$_{25}$          \\ \midrule
			\checkmark            &             &                   & 41.1        & 64.6          & \textbf{79.7} \\
			\checkmark            & \checkmark           &                   & 45.7        & \textbf{68.4} & 79.5          \\
			\checkmark            & \checkmark           & \checkmark                 & \textbf{46.0} & 67.6          & 78.9          \\ \bottomrule
		\end{tabular}
		\caption{The impact of each branch in top-down refinement on ScanNet v2 validation set.}
		\label{tab:top_down_refinement}
	\end{table}
	
	\begin{table}[]
		\centering
		\begin{tabular}{lccc}
			\toprule
			Category from class branch? & AP          &AP$_{50}$ & AP$_{25}$          \\ \midrule
			N                            & 45.0 & 65.6 & 76.2 \\
			Y                            & \textbf{46.0} & \textbf{67.6} & \textbf{78.9} \\ \bottomrule
		\end{tabular}
		\caption{Ablation study on instance category. ``N" indicates that the instance category is taken from majority vote of semantic prediction. ``Y" indicates that the instance category is taken from classification branch}
		\label{tab:instance_category}
	\end{table}
	
	\paragraph{Top-Down Refinement.}
	We further provide the ablation results on the top-down refinement, on \Cref{tab:top_down_refinement}. With only the classification branch, our method achieves AP/AP$_{50}$/AP$_{25}$ of 41.1/64.6/79.7(\%). When mask branch and mask scoring branch are in turn applied, the performance tends to improve on the higher IoU threshold regions. Combining all branches yields the performance  AP/AP$_{50}$/AP$_{25}$ of 46.0/67.6/78.9(\%).
	
	\paragraph{Instance Category from Classification Branch.}
	\Cref{tab:instance_category} reports the results of different schemes to obtain object categories. The results show that deriving the object category from semantic prediction yields the AP/AP$_{50}$/AP$_{25}$ of 45.0/65.6/76.2(\%). The proposed method directly uses the output of the classification branch as the instance class. The classification branch aggregates all point features of the instance and classifies the instance with a single label, leading to more reliable prediction. The results show that directly using classification output as object category improves the AP/AP$_{50}$/AP$_{25}$ to 46.0/67.6/78.9(\%).
	
	\section{Conclusion}
	We have presented SoftGroup, a simple yet effective method for instance segmentation on 3D point clouds. SoftGroup performs grouping on soft semantic scores to address the problem stemming from hard grouping on locally ambiguous objects. The instance proposals obtained from the grouping stage are assigned to either positive or negative samples. Then a top-down refinement stage is constructed to refine the positives and suppress the negatives. Extensive experiments on different datasets show that our method outperforms the existing state-of-the-art method by a significant margin of +6.2\% on the hidden ScanNet v2 test set and +6.8\% on S3DIS Area 5 in terms of AP$_{50}$. SoftGroup is also fast, requiring 345ms to process a ScanNet scene.
	
	{\small
		\bibliographystyle{ieee_fullname}
		\bibliography{refs}
	}
	
\end{document}